\documentclass[12pt]{article}
\usepackage{amsmath,amsthm}
\usepackage{bm,amssymb,mathtools}
\usepackage[ruled,vlined]{algorithm2e}
\usepackage{graphicx,psfrag,epsf}
\usepackage{enumerate}
\usepackage[round,authoryear]{natbib}
\usepackage{url} 
\usepackage[colorlinks,citecolor=blue,urlcolor=blue,linkcolor = blue,backref=page]{hyperref}
\usepackage{float}
\usepackage{siunitx}
\usepackage{multirow}
\usepackage{booktabs}
\usepackage{tikz}
\usepackage{subcaption}
\usetikzlibrary{positioning,arrows.meta,shapes.geometric,calc,decorations.pathreplacing}

\title{Neural Architectures for Amortized Bayesian Inference: Statistical Foundations and Empirical Assessments}
\author{Roy Shivam Ram Shreshtth$^1$, Arnab Hazra$^1$, and Gourab Mukherjee$^2$\\\\
$^1$Department of Statistics and Data Science, \\ Indian Institute of Technology Kanpur, Kanpur, India 208016\\
$^2$Marshall School of Business, \\ University of Southern California, Los Angeles, United States 90089-0809}

\date{}
\begin{document}

\maketitle

\begin{abstract}
\noindent 
Since the turn of the century, approximate Bayesian inference has steadily evolved as new computational techniques have been incorporated to handle increasingly complex, large-scale predictive problems. The recent success of deep neural networks and foundation models has now given rise to a new paradigm in statistical modeling, in which Bayesian inference can be amortized through large-scale learned predictors. In amortized inference, substantial computation is required at the beginning to train a neural network, but it can subsequently produce approximate posteriors or predictions at much lower computational cost across a wide range of tasks. While the typical Bayesian inference procedures are computationally expensive due to repeated likelihood calculations and Monte Carlo steps for each new dataset, amortized inference provides a much lower computational cost at deployment. 

Despite the growing popularity of amortized inference, its statistical interpretation and position within Bayesian inference remain poorly explored. In this paper, we present a statistical perspective on several major neural architectures, including feedforward networks, Deep Sets, and Transformers, and examine how they naturally support amortized Bayesian inference. We explore how these models perform structured approximation and also probabilistic reasoning in ways that yield controlled generalization error throughout a wide range of deployment scenarios, and how these properties can be harnessed for Bayesian computation. Via simulation studies, we evaluate the accuracy, robustness, and uncertainty quantification of amortized inference across varying sample sizes, varying noise distributional families, varying sparsity levels, and multimodality, highlighting its strengths and limitations.
\end{abstract}


\section{Introduction}

Nearly twenty-five years ago, \citet{breiman2001statistical} drew attention to a divide within statistical methodology between a data modeling culture, rooted in explicit probabilistic formulations, and an algorithmic modeling culture, which prioritizes predictive performance through flexible procedures that often lack a generative interpretation. While the former traditionally guided Bayesian analysis, the past two decades have seen a rapid expansion of high-dimensional and highly nonlinear models arising from machine learning, which increasingly influence Bayesian workflows and expand the set of problems for which fully probabilistic treatment is computationally feasible \citep{shmueli2010explain}.

A major force behind this shift is the development of scalable Bayesian computation. Classical tools such as Markov chain Monte Carlo (MCMC) provide well-understood inferential guarantees \citep{gelfand1990sampling}. Still, they can become computationally prohibitive when applied repeatedly across large collections of related datasets or in models with expensive likelihoods. In many modern scientific pipelines, including climate modeling, astrophysics, genomics, and simulation-based inference, analysts must solve thousands or millions of Bayesian inverse problems within tight computational budgets. In such settings, the traditional per-instance paradigm of Bayesian computation becomes infeasible.

The concept of amortized Bayesian inference addresses this challenge by shifting the computational burden from inference at deployment time to a substantial, upfront training phase \citep{gershman2014amortized}. Instead of running MCMC or optimizing a variational objective separately for each dataset, a flexible neural network is trained on a large collection of simulated or historical Bayesian inference tasks. Once trained, this network provides approximate posterior summaries, including point estimates, densities, or samples, for new datasets at negligible marginal cost. In effect, amortization constructs a global Bayesian surrogate, enabling rapid inference whenever new data arise from the same or a closely related generative mechanism. The benefits of amortization are well-illustrated by the training-deployment dynamics of modern large language models. For example, the BigScience Large Open-Science Open-access Multilingual Language Model (BLOOM) required more than a million GPU-hours on several hundred high-end processors during its four-month training phase, consuming an estimated 433.2 MWh of energy \citep{luccioni2023estimating}. As pointed out by \cite{zammit2025neural}, this amount of energy is roughly equivalent to the annual usage of 70 Australian households. Once this substantial upfront cost is incurred, the trained model can generate high-quality text on a single GPU at a negligible computational and energy cost. This stark disparity between the cost of training and the efficiency of subsequent inference epitomizes the principle of amortization: a large, one-time computational investment yields an inference mechanism that can be applied repeatedly and efficiently across new inputs.

Amortized Bayesian inference has recently emerged as a powerful alternative to likelihood-based inference for complex data structures, including complex spatial and extreme-value models, where exact likelihoods are intractable or prohibitively expensive \citep{sainsbury2024likelihood, zammit2025neural}. In spatial statistics, \cite{sainsbury2025neural} have shown that neural Bayes estimators and graph neural-network–based amortized inferences can deliver fast, accurate inference for Gaussian and non-Gaussian spatial fields, irregularly spaced data, and high-dimensional latent structures, dramatically reducing computation relative to classical Monte Carlo or Integrated Nested Laplace Approximation \citep{rue2009approximate} approaches. In extreme-value analysis, where censored peaks-over-threshold models, heavy-tailed dependence, and spatiotemporal extremes yield challenging likelihood surfaces, amortized Bayesian inference has been employed to construct likelihood-free estimators, accelerate GEV-type Bayesian workflows, and facilitate inference under censoring and weak identifiability \citep{richards2024neural}. Collectively, these developments demonstrate that amortized Bayesian inference provides a scalable and flexible framework for tackling the computational bottlenecks inherent in modern spatial and extreme-value modeling, offering near-instant, reusable inference once a simulator-trained network has been fit.

Neural amortization strategies can be broadly classified according to the type of Bayesian quantity they approximate. The first regime, direct neural point estimation, learns a deterministic mapping from data to parameter estimates. Although limited to first-order summaries, this approach is useful in likelihood-free or simulation-based settings where exact Bayesian computation is challenging. Architectures such as Deep Sets \citep{zaheer2017deepsets} and Set Transformers \citep{lee2019set} provide principled methods for encoding symmetries inherent in the likelihood or prior structure. Neural posterior estimation (NPE) extends this idea by learning conditional density estimators for the posterior \citep{greenberg2019automatic}. The second regime, amortized variational inference (AVI), provides approximate posterior distributions rather than point estimates. These methods train neural networks to output parameters of an approximate posterior, effectively learning a variational family indexed by the data. Modern formulations often employ normalizing flows \citep{rezende2015variational, papamakarios2021normalizing} to represent posteriors with complex curvature, skewness, and multimodality that cannot be captured by mean-field approximations \citep{radev2024neural}. In Bayesian computation, AVI can be viewed as a simultaneous variational approximation over an entire distribution of tasks, rather than a single dataset. The third regime, neural samplers and transport methods, aims to accelerate Bayesian sampling by learning transformations that map complex posterior distributions to simpler latent distributions, where standard Monte Carlo procedures can mix rapidly. Neural transport maps \citep{hoffman2019neutra} and flow-matching methods \citep{lipman2023flow} exemplify this approach. These techniques serve as learned reparameterizations or preconditioners for MCMC, yielding substantial computational savings for models with strongly correlated parameters or irregular posterior geometries.

The purpose of this paper is two-fold. We begin by reviewing the statistical background of a few machine learning tools and concepts specific to amortized Bayesian inference. Subsequently, we assess these amortized approaches from a Bayesian perspective, with particular attention to their reliability, structural limitations, and empirical performance relative to classical computational methods. While amortization can offer significant gains in speed when the training distribution accurately represents the deployment environment, its accuracy and stability degrade once the generative mechanism deviates from the conditions encountered during training. To study this behavior systematically, we conduct extensive simulation experiments along four axes. First, we investigate training heterogeneity, examining how variations in task size and diversity influence posterior accuracy and calibration. Second, we study data-scarce deployment and quantify the dependence of amortized posteriors on effective sample size. Third, we evaluate robustness under noise distributional mismatch, characterizing how predictive and posterior accuracy deteriorate as covariates or structural aspects of the data-generating process drift from the training regime. Fourth, we examine whether amortized neural inference can accurately capture posterior distributions with disconnected support and complex topology, where conventional unimodal approximations are inadequate.

The paper proceeds as follows. Section~\ref{sec:architecture} reviews the neural architectures employed: Feedforward Networks, Deep Sets, and Transformers, and clarifies their connection to Bayesian computation through kernel, variational, and transport-theoretic interpretations. Different aspects of amortized inference are summarized in Section~\ref{sec:amortized}. Section \ref{sec:simulation} presents simulation studies examining the four regimes described above under varying sample sizes, varying noise distributional families, varying sparsity levels, and multimodality.\footnote{All code used to generate the results is provided at \url{https://github.com/Royshivam18/Neural-Amortized-Inference} to ensure full reproducibility.} Section \ref{sec:discussion} summarizes the findings and highlights some future research directions.

\section{Popular Neural Network Architectures}
\label{sec:architecture}

In this section, we discuss modern neural network architectures commonly used in amortized inference pipelines. By drawing analogies to classical statistical methods, we explain why each architecture is suited to particular inference tasks rather than treating them as black-box function approximators.

\paragraph{Notation and conventions.}
We write vectors as columns by default; the only exception is Section~\ref{sec:transformer}, where the input matrix $X \in \mathbb{R}^{N \times d_{\mathrm{model}}}$ stacks token embeddings as rows. Throughout the paper, $\theta$ denotes the trainable weights of the amortized estimator and $\vartheta \in \Theta$ the inferential target (a generic Bayesian parameter). The symbol $\sigma$ is used for both the component-wise nonlinear activation in an FNN (Section~\ref{sec:fnn-basis}) and standard deviations elsewhere--the meaning is clear from context. By a \emph{token} we mean an indexed unit of an input set; for tabular regression, a token is one observation $(x_n, y_n)$. An \emph{epoch} is one pass over the meta-training set of tasks, and the per-task sample size $N_t$ is always distinguished from the meta-training-set size $T_{\mathrm{train}}$.

\subsection{Feedforward Neural Networks}

The Feedforward Neural Network (FNN), also known as a multilayer perceptron (MLP) \citep{rumelhart1986learning}, is the foundational architecture for deep learning. The Universal Approximation Theorem (recalled in Section~\ref{sec:fnn-basis}) guarantees that FNNs can represent arbitrarily complex nonlinear mappings between inputs and outputs.

The mathematical setup of FNN is as follows. Let $L \in \mathbb{N}$ denote the network depth. We define the width of each layer as $n_l$, where $l \in \{0, \dots, L\}$. Consistent with our notation, the input dimension is $n_0 = d_{\mathrm{in}}$ and the output dimension is $n_L = d_{\mathrm{out}}$. A (dense) feedforward neural network is a parametrized mapping $f_\theta: \mathbb{R}^{d_{\mathrm{in}}} \longrightarrow \mathbb{R}^{d_{\mathrm{out}}}$, where $\theta = \bigl\{(W^{[l]}, b^{[l]})\bigr\}_{l=1}^{L}$, constructed on a strictly layered, acyclic computational graph with full connectivity between consecutive layers. Given an input vector $x \in \mathbb{R}^{d_{\mathrm{in}}}$, the network computes its output through a sequence of recursive transformations. The computation begins by setting the initial activation equal to the input, $a^{[0]} = x$. For each hidden layer \(l = 1, \dots, L-1\), the network forms an affine transformation of the activations of the previous layer as
\begin{equation}
\label{eq:affine_fnn}
    z^{[l]} = W^{[l]}a^{[l-1]} + b^{[l]},
\end{equation}
and then applies a nonlinear activation function $\sigma^{[l]}$ component-wise to obtain
\begin{equation}
\label{eq:activation}
    a^{[l]} = \sigma^{[l]}(z^{[l]}).
\end{equation}
Here, $W^{[l]} \in \mathbb{R}^{n_l \times n_{l-1}}$ and $b^{[l]} \in \mathbb{R}^{n_l}$  in \eqref{eq:affine_fnn} denote the weight matrix and bias vector for layer $l$, respectively. An affine transformation of the last hidden-layer activation produces the final output of the network:
$$f_{\theta}(x) = W^{[L]}a^{[L-1]} + b^{[L]}$$
typically without an additional nonlinearity.

The nonlinear activation functions $\sigma^{[l]}$ in \eqref{eq:activation} are essential. For instance, the Rectified Linear Unit (ReLU), defined as $\sigma(z) = \max(0, z)$, restricts the network to piecewise-linear functions. Without these nonlinearities, the composition collapses to a single affine map $f_{\theta}(x) = W_{\mathrm{total}}x + b_{\mathrm{total}}$, and expressivity is lost.

The complete parameter set $\theta$ resides in the space $\Theta = \prod_{l=1}^{L} ( \mathbb{R}^{n_l \times n_{l-1}} \times \mathbb{R}^{n_l} )$, and the total number of scalar parameters is given by $N_\theta = \sum_{l=1}^{L} n_l(n_{l-1} + 1)$. This hierarchical composition of functions allows the FNN to construct highly complex decision boundaries from simple linear operations. Figure~\ref{fig:fnn_arch} summarizes this structure schematically.

\begin{figure}[t]
\centering
\begin{tikzpicture}[
  font=\footnotesize,
  >={Latex[length=2mm,width=1.6mm]},
  neuron/.style={circle, draw=black!75, line width=0.5pt, minimum size=6mm, inner sep=0pt, fill=white},
  inp/.style={neuron, fill=gray!10},
  hid/.style={neuron, fill=blue!8},
  basis/.style={neuron, fill=orange!30, draw=orange!75!black, line width=0.9pt},
  out/.style={neuron, fill=gray!10},
  layerlbl/.style={align=center, text width=2.7cm, font=\small, anchor=north},
  wlbl/.style={align=center, font=\footnotesize, anchor=south},
  arr/.style={->, gray!45, line width=0.3pt},
  arrlast/.style={->, orange!65!black, line width=0.5pt}
]

\def\xI{0}
\def\xA{2.8}
\def\xB{5.4}
\def\xDots{6.9}
\def\xP{8.6}
\def\xO{11.4}

\foreach \y [count=\i] in {1.4, 0.5, -0.4, -1.3}
  \node[inp] (I\i) at (\xI, \y) {};
\node[layerlbl] at (\xI, -2.0) {Input \\ $x \in \mathbb{R}^{d_{\mathrm{in}}}$};

\foreach \y [count=\i] in {1.8, 0.9, 0.0, -0.9, -1.8}
  \node[hid] (A\i) at (\xA, \y) {};
\node[layerlbl] at (\xA, -2.4) {$a^{[1]} = \sigma^{[1]}(z^{[1]})$};

\foreach \y [count=\i] in {1.8, 0.9, 0.0, -0.9, -1.8}
  \node[hid] (B\i) at (\xB, \y) {};
\node[layerlbl] at (\xB, -2.4) {$a^{[2]} = \sigma^{[2]}(z^{[2]})$};

\node[font=\Large] at (\xDots, 0) {$\cdots$};

\foreach \y [count=\i] in {1.8, 0.9, 0.0, -0.9, -1.8}
  \node[basis] (P\i) at (\xP, \y) {};
\node[layerlbl, text=orange!75!black] at (\xP, -2.4) {Basis layer \\ $\Phi(x) = a^{[L-1]}$};

\foreach \y [count=\i] in {0.45, -0.45}
\node[circle,fill=black,inner sep=1pt] (O\i) at (\xO,\y) {};
\node[layerlbl] at (\xO, -2.0) {Output \\ $f_\theta(x) \in \mathbb{R}^{d_{\mathrm{out}}}$};

\foreach \i in {1,...,4} \foreach \j in {1,...,5} \draw[arr] (I\i) -- (A\j);
\foreach \i in {1,...,5} \foreach \j in {1,...,5} \draw[arr] (A\i) -- (B\j);
\foreach \i in {1,...,5} \draw[arr, dashed] (B\i) -- ($(\xDots,0) + (-0.45,0)$);
\foreach \j in {1,...,5} \draw[arr, dashed] ($(\xDots,0) + (0.45,0)$) -- (P\j);
\foreach \i in {1,...,5} \foreach \j in {1,2} \draw[arrlast] (P\i) -- (O\j);

\node[wlbl] at ({(\xI+\xA)/2}, 2.4) {$W^{[1]},\ b^{[1]}$};
\node[wlbl] at ({(\xA+\xB)/2}, 2.4) {$W^{[2]},\ b^{[2]}$};
\node[wlbl] at ({(\xP+\xO)/2}, 2.4) {$W^{[L]},\ b^{[L]}$};

\node[anchor=north, font=\scriptsize, text=black!60] at ({(\xI+\xA)/2}, 2.5) {affine, then $\sigma^{[1]}$};
\node[anchor=north, font=\scriptsize, text=black!60] at ({(\xA+\xB)/2}, 2.5) {affine, then $\sigma^{[2]}$};
\node[anchor=north, font=\scriptsize, text=black!60] at ({(\xP+\xO)/2}, 2.5) {affine (linear read-out)};

\end{tikzpicture}
\caption{Architecture of an $L$-layer feedforward neural network. Each hidden layer applies an affine map $z^{[l]} = W^{[l]} a^{[l-1]} + b^{[l]}$ followed by the component-wise nonlinearity $\sigma^{[l]}$; the output layer is a linear read-out $f_\theta(x) = W^{[L]} a^{[L-1]} + b^{[L]}$. The penultimate layer $a^{[L-1]} = \Phi(x)$, shown in orange, is highlighted as a learned dictionary of \emph{basis functions}: the network output is a linear combination of the entries of $\Phi(x)$, so the FNN realizes an adaptive basis expansion (Section~\ref{sec:fnn-basis}). In the Transformer of Section~\ref{sec:transformer}, the query, key, and value projections are themselves FNN outputs of exactly this form, and the layer $\Phi(x)$ corresponds to the features on which attention then operates.}
\label{fig:fnn_arch}
\end{figure}

\subsubsection{Universal Approximation and Adaptive Basis Functions}
\label{sec:fnn-basis}

The utility of FNNs in statistical inference is rigorously grounded in the Universal Approximation Theorem. Seminal works by \citet{cybenko1989approximation} and \citet{hornik1991approximation} established that standard multilayer feedforward networks are universal approximators. Formally, let $C(K)$ denote the space of continuous functions on a compact set $K \subset \mathbb{R}^{d_{\mathrm{in}}}$. The theorem states that the set of functions representable by a neural network with at least one hidden layer and a non-polynomial activation function $\sigma$ is dense in $C(K)$ with respect to the supremum norm. That is, for any target function $g \in C(K)$ and error tolerance $\epsilon > 0$, there exists a parameter configuration $\theta$ such that
\[
\sup_{x \in K} |g(x) - f_\theta(x)| < \epsilon.
\]
While the theorem guarantees the existence of an approximating network, the geometric nature of the approximation is determined by the choice of activation function.

Modern architectures predominantly use ReLU activations. As shown by \citet{montufar2014number} and \citet{arora2018understanding}, a ReLU network is a piecewise-affine estimator: it partitions the input domain into finitely many convex polytopes and is affine on each. Nonlinear targets are thus approximated by tiling them with flat facets, analogous to a polygonal mesh approximating a curved surface. In contrast, classical activation functions like the logistic sigmoid, $\sigma(z) = (1+e^{-z})^{-1}$, or the hyperbolic tangent, $\sigma(z) = \tanh(z)$, operate via smooth superposition. As established in the convergence rates derived by \citet{barron1993universal}, networks using these smooth, bounded, sigmoidal functions approximate the target function by superimposing smooth `steps' or `ridges'. Unlike the sharp boundaries of ReLU polytopes, these activations provide a differentiable approximation where nonlinearity arises from the smooth saturation of neurons \citep{cybenko1989approximation}.


We can regard the first hidden layer as learning a dictionary of feature maps denoted by $\Phi(x)= (\phi_1(x), \dots, \phi_{n_1}(x))^\top$, where each $\phi_i(x) := \sigma((W_i^{[1]})^\top x + b_i^{[1]})$ is a nonlinear basis function. The second hidden layer then computes weighted affine combinations of these features. Analytically, the network output can be written as a basis expansion given by
\[
f_{\theta}(x) \;=\; c + \sum_{j=1}^{n_2} \beta_j \sigma \bigl( (W_j^{[2]})^\top \Phi(x) + b_j^{[2]} \bigr),
\]
where $W_j^{[2]}$ denotes the $j$-th row of the weight matrix $W^{[2]}$. This expression clarifies that the network is learning a new family of basis functions $\{\sigma((W_j^{[2]})^\top \Phi(x) + b_j^{[2]})\}_{j=1}^{n_2}$ built upon the first-layer dictionary, rather than using a fixed basis. Unlike classical nonparametric regression, where the basis (e.g., Fourier or spline) is fixed \textit{a priori}, the FNN adaptively learns the basis parameters to represent the data structure best. This adaptive capability allows the FNN to emulate classical bases. Let $\mathcal{G} = \{g_1, \dots, g_p\} \subset C(K)$ be any finite collection of continuous functions (a prescribed basis). By the universal approximation theorem, for any $\epsilon > 0$, there exist parameters such that
\[
\sup_{x \in K} \Bigl| f_{\theta}(x) - \sum_{j=1}^{p} \beta_j g_j(x) \Bigr| \;<\; \epsilon.
\]

Equivalently, there exists a partition of the hidden units into blocks $\{\mathcal{I}_j\}_{j=1}^p$ such that each block realizes an approximation:
\[
\tilde{g}_j(x) \;=\; \sum_{i \in \mathcal{I}_j} \alpha_i^{(j)} \sigma((W_i^{[1]})^\top x + b_i^{[1]}),
\quad \sup_{x \in K} |g_j(x) - \tilde{g}_j(x)| < \varepsilon/p.
\]
The second layer learns coefficients to form $\sum_j \beta_j \tilde{g}_j(x)$. Hence, even if the first layer (dictionary) is fixed, a two-hidden-layer network can approximate any finite linear combination of target basis functions on $K$.

This perspective allows us to map FNNs directly to classical kernel methods. To see this connection, we shift our focus from a trained deep network to a single-hidden-layer network with randomized first-layer parameters:
\[
W_i^{[1]} \sim P_W, \qquad b_i^{[1]} \sim P_b.
\]
If we freeze these parameters at initialization and only train the output-layer weights, each hidden unit activation $\phi_i(x) = \sigma((W_i^{[1]})^\top x + b_i^{[1]})$ acts strictly as a random feature \citep{rahimi2007random}. The induced kernel$$k(x, x^{\prime}) \;:=\; \mathbb{E}_{W, b} \left[\sigma((W^{[1]})^\top x + b^{[1]}) \sigma((W^{[1]})^\top x^{\prime} + b^{[1]})\right]$$is well-defined under $P_W$ and $P_b$. Training the output coefficients $\{\beta_j\}_{j=1}^{n_1}$ with squared loss and an $\ell_2$-penalty is equivalent to kernel ridge regression with kernel $k(\cdot, \cdot)$. Moreover, \citet{neal1996priors} showed that if we place independent Gaussian priors on these output weights and let the hidden layer width $n_1 \to \infty$ with appropriate scaling, the induced prior over functions converges to a Gaussian Process (GP) with covariance $k(\cdot, \cdot)$. In this regime, the posterior mean predictor coincides with the GP regression estimator under kernel $k(\cdot, \cdot)$.

This basis-learning behavior is observable empirically. On algorithmic tasks such as modular addition ($a + b \mod P$), networks initially memorize the training data and then, after extended training, transition to perfect generalization--a phenomenon known as \emph{grokking} \citep{power2022grokking}. In the generalizing phase, the embedding layers converge to Fourier-like bases (sines and cosines) over the cyclic group \citep{nanda2023progress}, an empirical confirmation that the network discovers a problem-appropriate basis rather than fitting a fixed dictionary.

\subsection{Deep Sets as the Summary Statistic Machine}

Standard FNNs are inherently limited when applied to statistical inference tasks involving sets of independent and identically distributed (IID) observations. A standard MLP requires a fixed-size input vector, yet datasets in inference tasks naturally possess variable cardinality. Furthermore, the joint probability of an IID set, $p(x_1, \dots, x_N)$, is invariant to permutations of the indices. Consequently, any estimator $\widehat{\theta}(x_1, \dots, x_N)$ derived from such data must respect this exchangeability. Standard architectures, which process inputs as ordered sequences, fail to capture this symmetry efficiently. To address this, \citet{zaheer2017deepsets} proposed the `Deep Sets' architecture, which provides a theoretically grounded framework for learning permutation-invariant functions.

Consider a set of inputs $\mathcal{X} = \{x_1, x_2, \dots, x_N\}$ where each element $x_j \in \mathbb{R}^{d_{\mathrm{in}}}$. The Deep Sets architecture processes this set through a three-stage mechanism to ensure permutation invariance. First, an embedding function (encoder) $\phi: \mathbb{R}^{d_{\mathrm{in}}} \to \mathbb{R}^{d_{\mathrm{latent}}}$, typically parameterized as an MLP, maps each individual element to a latent representation:
\begin{equation}
\nonumber    h_j = \phi(x_j), \quad j=1, \dots, N.
\end{equation}
Crucially, this mapping is applied identically to every element in the set, ensuring that feature extraction is independent of the position of the element.

Subsequently, a permutation-invariant aggregation operator is applied to the set of latent codes $\{h_1, \dots, h_N\}$. While operators such as the mean or maximum are permissible, sum-pooling is often preferred in theoretical contexts as it preserves information regarding the set size $N$, which is vital for Bayesian inference. The aggregated summary statistic $Z$ is computed as
\begin{equation}
\nonumber    Z = \sum_{j=1}^N h_j = \sum_{j=1}^N \phi(x_j).
\end{equation}
Finally, a second neural network $\rho: \mathbb{R}^{d_{\mathrm{latent}}} \to \mathbb{R}^{d_{\mathrm{out}}}$, referred to as the decoder or regressor, processes the aggregated summary statistic to produce the final estimator
\begin{equation}
\nonumber    \widehat{y} = \rho(Z).
\end{equation}
Combining these steps, the complete Deep Sets estimator can be expressed in the compact form
\begin{equation}
 f(\mathcal{X}) \;=\; \rho\left( \sum_{x \in \mathcal{X}} \phi(x) \right).
\end{equation}
Figure~\ref{fig:deepsets_arch} depicts this three-stage construction.

\begin{figure}[t]
\centering
\begin{tikzpicture}[
  font=\footnotesize,
  >={Latex[length=2mm,width=1.6mm]},
  point/.style={circle, draw=black!75, line width=0.5pt, minimum size=4.5mm, inner sep=0pt, fill=gray!10},
  encbox/.style={rectangle, draw=black!75, line width=0.6pt, rounded corners=2pt, fill=blue!8, minimum width=1.5cm, minimum height=0.75cm, align=center},
  hvec/.style={rectangle, draw=black!75, line width=0.5pt, fill=blue!8, minimum width=0.85cm, minimum height=0.45cm, align=center, inner sep=2pt},
  sumbox/.style={rectangle, draw=orange!75!black, line width=1.0pt, rounded corners=3pt, fill=orange!28, minimum width=2.0cm, minimum height=1.4cm, align=center},
  decbox/.style={rectangle, draw=black!75, line width=0.6pt, rounded corners=2pt, fill=blue!8, minimum width=1.5cm, minimum height=0.75cm, align=center},
  outv/.style={rectangle, draw=black!75, line width=0.5pt, fill=gray!10, minimum width=0.9cm, minimum height=0.55cm, align=center, inner sep=2pt},
  arr/.style={->, gray!55, line width=0.4pt},
  arrhi/.style={->, orange!70!black, line width=0.55pt},
  layerlbl/.style={align=center, font=\small, anchor=north}
]

\def\xX{-1.0}
\def\xPhi{2.0}
\def\xH{3.6}
\def\xZ{6.6}
\def\xRho{9.2}
\def\xY{11}

\node[point] (x1) at (\xX, 1.6)  {};  \node[anchor=east, font=\scriptsize] at ($(x1.west)+(-0.05,0)$) {$x_1$};
\node[point] (x2) at (\xX, 0.7)  {};  \node[anchor=east, font=\scriptsize] at ($(x2.west)+(-0.05,0)$) {$x_2$};
\node[point] (x3) at (\xX, -0.2) {};  \node[anchor=east, font=\scriptsize] at ($(x3.west)+(-0.05,0)$) {$x_3$};
\node[font=\large] at (\xX, -1.0) {$\vdots$};
\node[point] (xN) at (\xX, -1.8) {};  \node[anchor=east, font=\scriptsize] at ($(xN.west)+(-0.05,0)$) {$x_N$};
\node[layerlbl, text width=2.7cm] at (\xX, -2.4) {Unordered set \\ $\mathcal{X}=\{x_1,\dots,x_N\}$};

\node[encbox] (phi1) at (\xPhi, 1.6)  {$\phi(\cdot)$};
\node[encbox] (phi2) at (\xPhi, 0.7)  {$\phi(\cdot)$};
\node[encbox] (phi3) at (\xPhi, -0.2) {$\phi(\cdot)$};
\node[encbox] (phiN) at (\xPhi, -1.8) {$\phi(\cdot)$};
\node[layerlbl, text width=2.8cm] at (\xPhi, -2.4) {Shared encoder \\ $\phi:\mathbb{R}^{d_{\mathrm{in}}}\!\to\!\mathbb{R}^{d_{\mathrm{lat}}}$};
\draw[decorate, decoration={brace, mirror, amplitude=4pt}, gray!55] ($(phi1.east)+(0.08,0)$) -- ($(phiN.east)+(0.08,0)$);
\node[anchor=west, font=\scriptsize, text=gray!50!black, align=center, text width=1.4cm] at ($(phi3.east)+(0.9,0.05)$) {shared\\ weights};

\node[hvec] (h1) at (\xH, 1.6)  {$h_1$};
\node[hvec] (h2) at (\xH, 0.7)  {$h_2$};
\node[hvec] (h3) at (\xH, -0.2) {$h_3$};
\node[hvec] (hN) at (\xH, -1.8) {$h_N$};

\node[sumbox] (Z) at (\xZ, -0.1) {$\displaystyle Z = \sum_{j=1}^{N} \phi(x_j)$};
\node[layerlbl, text=orange!70!black, text width=3.2cm] at (\xZ, -2.4) {Aggregation: \\ empirical generalised moments};

\node[decbox] (rho) at (\xRho, -0.1) {$\rho(\cdot)$};
\node[layerlbl, text width=2.8cm] at (\xRho, -2.4) {Decoder \\ (inverse-moment map)};

\node[outv] (y) at (\xY, -0.1) {$\widehat{y}$};
\node[layerlbl, text width=2.0cm] at (\xY, -2.4) {Output};

\draw[arr] (x1) -- (phi1);
\draw[arr] (x2) -- (phi2);
\draw[arr] (x3) -- (phi3);
\draw[arr] (xN) -- (phiN);

\draw[arr] (phi1) -- (h1);
\draw[arr] (phi2) -- (h2);
\draw[arr] (phi3) -- (h3);
\draw[arr] (phiN) -- (hN);

\draw[arrhi] (h1) -- (Z.west);
\draw[arrhi] (h2) -- (Z.west);
\draw[arrhi] (h3) -- (Z.west);
\draw[arrhi] (hN) -- (Z.west);

\draw[arr] (Z.east) -- (rho.west);
\draw[arr] (rho.east) -- (y.west);

\end{tikzpicture}
\caption{Deep Sets architecture. Each element $x_j$ of an unordered set $\mathcal{X}=\{x_1,\dots,x_N\}$ is passed through the same encoder $\phi$ (shared parameters), producing latent codes $h_j = \phi(x_j)$. The codes are aggregated by a permutation-invariant sum, $Z = \sum_{j=1}^{N} \phi(x_j)$, shown in orange. This aggregation block acts as a vector of \emph{empirical generalized moments}: it is the neural analog of the classical method-of-moments statistic $\hat m_k = \tfrac{1}{N}\sum_j x_j^k$, except that the fixed polynomial basis is replaced by the learned features $\phi$. The decoder $\rho$ then approximates the inverse moment-to-parameter map and returns $\widehat{y} = \rho(Z)$.}
\label{fig:deepsets_arch}
\end{figure}

From a statistical perspective, this architecture acts as a neural generalization of the Method of Moments (MoM). In classical estimation, parameters $\theta$ are estimated by equating empirical moments to theoretical moments. This estimation involves computing a vector of sample moments, $\hat{m}_k = \frac{1}{N} \sum x_j^k$, and applying an inverse mapping $g$ to solve for $\theta$. The Deep Sets formulation mirrors this structure exactly: the encoder $\phi(x)$ learns to extract optimal `generalized moments' (sufficient statistics) rather than relying on fixed polynomial powers; the summation $\sum \phi(x_j)$ accumulates these statistics into a fixed-size summary; and the outer network $\rho$ approximates the inverse mapping $g$.

This analogy extends to the Moment Generating Function (MGF). The Taylor expansion of the MGF, $M_X(t) = \mathbb{E}[e^{t^\top X}]$, encapsulates all moments of the distribution. If the encoder $\phi(x)$ possesses sufficient capacity, the aggregated vector $\sum \phi(x)$ effectively approximates the MGF of the underlying empirical distribution. Since the MGF uniquely characterizes the distribution under mild conditions, a Deep Set that learns a sufficient number of these moments captures a finite-dimensional representation of the entire probability measure \citep{wagstaff2019limitations}. Theoretically, this principle is grounded in the Fisher-Neyman Factorization Theorem, which states that for Exponential Families, the likelihood factorizes through a sufficient statistic $T(x)$. Deep Sets effectively learn this statistic via $\phi$, making them `Neural Sufficient Statistic Learners'.

\subsection{The Transformer as a Statistical Operator}
\label{sec:transformer}

Statistically, a Transformer layer can be read as a learned, data-dependent kernel smoother on the input set. The variant we use in the experiments is the Set Transformer of \citet{lee2019set}, which applies the same construction described below to an \emph{unordered} input set (no positional encoding), yielding a permutation-equivariant operator over samples. The kernel-smoothing interpretation developed in this subsection applies identically to both.

Let $X \in \mathbb{R}^{N \times d_{\mathrm{model}}}$ denote the input sequence of token embeddings, where the row $x_j \in \mathbb{R}^{d_{\mathrm{model}}}$ corresponds to the $j$-th token. (We use the term \emph{token} for an indexed unit of the input set; for tabular regression, a token is one observation $(x_n, y_n)$.) Each Transformer layer \citep{vaswani2017attention} projects every token through three learned feed-forward networks $\phi_Q, \phi_K, \phi_V : \mathbb{R}^{d_{\mathrm{model}}} \to \mathbb{R}^{d_k}$, yielding queries, keys, and values:
\begin{equation}
\nonumber q_j = \phi_Q(x_j), \quad k_j = \phi_K(x_j), \quad v_j = \phi_V(x_j),
\end{equation}
where $q_j, k_j, v_j$ form the $j$-th rows of $Q, K, V \in \mathbb{R}^{N \times d_k}$. In the original Transformer of \citet{vaswani2017attention}, each of $\phi_Q, \phi_K, \phi_V$ is a single linear layer (i.e., $\phi_Q(x_j) = x_j W_Q$ for a learned matrix $W_Q$, and similarly for $K, V$); modern variants frequently replace these with small multilayer FFNs to increase expressivity. The mechanism then computes a pairwise similarity matrix $S \in \mathbb{R}^{N \times N}$ with entries $S_{ij} = \phi_Q(x_i)\, \phi_K(x_j)^{\!\top}$, representing the unnormalized alignment between the $i$-th query and the $j$-th key. The scores are scaled by $1/\sqrt{d_k}$ and passed through a row-wise softmax to yield the attention matrix $A$,
\begin{equation}
\nonumber A_{ij} \;=\; \frac{\exp(S_{ij}/\sqrt{d_k})}{\sum_{\ell=1}^N \exp(S_{i\ell}/\sqrt{d_k})}.
\end{equation}
The layer output is $Y = AV$, i.e., $Y_i = \sum_{j=1}^N A_{ij} v_j$. In standard implementations, several such \emph{heads} (parallel attention computations with independent $\phi_Q, \phi_K, \phi_V$) are concatenated to capture diverse dependencies. Figure~\ref{fig:transformer_arch} summarises this pipeline and lists the two attention forms commonly used downstream.

\begin{figure}[t]
\centering
\begin{tikzpicture}[
  font=\footnotesize,
  >={Latex[length=2mm,width=1.6mm]},
  token/.style={rectangle, draw=black!75, rounded corners=2pt, fill=gray!10, minimum width=0.7cm, minimum height=0.42cm, inner sep=2pt},
  fnnbox/.style={rectangle, draw=black!75, line width=0.7pt, rounded corners=3pt, fill=blue!10, minimum width=1.5cm, minimum height=0.95cm, align=center},
  qkvvec/.style={rectangle, draw=black!75, fill=blue!8, minimum width=0.85cm, minimum height=0.5cm, align=center, inner sep=2pt},
  attnbox/.style={rectangle, draw=orange!75!black, line width=1.0pt, rounded corners=4pt, fill=orange!28, minimum width=2.3cm, minimum height=1.7cm, align=center},
  outv/.style={rectangle, draw=black!75, fill=gray!10, minimum width=0.85cm, minimum height=0.55cm, align=center, inner sep=2pt},
  arr/.style={->, gray!55, line width=0.4pt},
  arrhi/.style={->, orange!70!black, line width=0.55pt},
  formulabox/.style={rectangle, draw=black!55, line width=0.4pt, rounded corners=3pt, fill=gray!4, align=left, inner sep=6pt, text width=5.4cm},
  layerlbl/.style={align=center, font=\small, anchor=north}
]


\node[token] (x1) at (0, 1.5)  {$x_1$};
\node[token] (x2) at (0, 0.7)  {$x_2$};
\node[token] (x3) at (0, -0.1) {$x_3$};
\node[font=\small] at (0, -0.8) {$\vdots$};
\node[token] (xN) at (0, -1.5) {$x_N$};
\draw[decorate, decoration={brace, amplitude=4pt}, gray!55]
  ($(x1.north east)+(0.18,0.0)$) -- ($(xN.south east)+(0.18,0.0)$);
\node[layerlbl, text width=2.6cm] at (0, -2.1) {Input tokens \\ $X \in \mathbb{R}^{N \times d_{\mathrm{model}}}$};

\node[fnnbox] (phiQ) at (2.6, 1.45)  {$\phi_Q$ \\[-1pt] \scriptsize FFN};
\node[fnnbox] (phiK) at (2.6, 0.0)   {$\phi_K$ \\[-1pt] \scriptsize FFN};
\node[fnnbox] (phiV) at (2.6, -1.45) {$\phi_V$ \\[-1pt] \scriptsize FFN};

\node[qkvvec] (Q) at (4.45, 1.45)  {$Q$};
\node[qkvvec] (K) at (4.45, 0.0)   {$K$};
\node[qkvvec] (V) at (4.45, -1.45) {$V$};

\node[attnbox] (attn) at (6.5, 0.0) {Scaled \\ dot-product \\ attention};
\node[layerlbl, text=orange!70!black, text width=3.2cm] at (6.5, -2.1) {Attention \\ \scriptsize{(in-context kernel smoother)}};

\node[outv] (Y) at (8.5, 0.0) {$Y$};
\node[layerlbl, text width=2.4cm] at (8.5, -2.1) {Output \\ $Y \in \mathbb{R}^{N \times d_k}$};

\foreach \src in {x1,x2,x3,xN} {
  \draw[arr] (\src.east) -- ($(phiQ.west)+(0,0)$);
  \draw[arr] (\src.east) -- ($(phiK.west)+(0,0)$);
  \draw[arr] (\src.east) -- ($(phiV.west)+(0,0)$);
}

\draw[arr] (phiQ.east) -- (Q.west);
\draw[arr] (phiK.east) -- (K.west);
\draw[arr] (phiV.east) -- (V.west);

\draw[arrhi] (Q.east) -- (attn.west);
\draw[arrhi] (K.east) -- (attn.west);
\draw[arrhi] (V.east) -- (attn.west);

\draw[arr] (attn.east) -- (Y.west);


\draw[gray!40, line width=0.4pt, dashed] (9.6, 2.2) -- (9.6, -2.4);

\node[anchor=north west, font=\small\bfseries] at (9.8, 2.2) {Attention forms};

\node[formulabox, anchor=north west, text width=120] (fL) at (9.8, 1.6) {%
\textbf{Linear similarity:}\\[3pt]
$\displaystyle S_{ij} \;=\; \frac{\phi_Q(x_i)\,\phi_K(x_j)^{\!\top}}{\sqrt{d_k}}$
};

\node[formulabox, anchor=north west, text width=120] (fS) at (9.8, -0.05) {%
\textbf{Softmax (normalized) \newline attention:}\\[3pt]
$\displaystyle A_{ij} \;=\; \frac{\exp\!\left(S_{ij}\right)}{\sum_{\ell=1}^{N}\exp\!\left(S_{i\ell}\right)},
\newline
\quad Y_i \;=\; \sum_{j=1}^{N} A_{ij}\,v_j$
};

\end{tikzpicture}
\caption{Transformer block. Every token $x_j$ is fed through three independent feed-forward networks $\phi_Q, \phi_K, \phi_V$--not bare matrix multiplications--producing per-token queries $q_j = \phi_Q(x_j)$, keys $k_j = \phi_K(x_j)$, and values $v_j = \phi_V(x_j)$. The scaled dot-product attention block (orange) aggregates value vectors using a data-dependent kernel induced by the query and key networks. The right panel lists the two attention strategies referenced in this section: the linear (unnormalized) similarity score $S_{ij}$, and the standard softmax-normalized attention $A_{ij}$ used to mix the values into the output $Y_i$.}
\label{fig:transformer_arch}
\end{figure}

The attention matrix $A$ admits a direct interpretation as a Nadaraya–Watson kernel regression estimator \citep{nadaraya1964estimating, watson1964smooth}. Define a data-dependent kernel $\kappa_\eta$ parameterized by the projection networks $\eta = \{\phi_Q, \phi_K\}$ as
\begin{equation}
\nonumber \kappa_\eta(x_i, x_j) \;=\; \exp\left(\frac{\phi_Q(x_i)\, \phi_K(x_j)^{\!\top}}{\sqrt{d_k}}\right).
\end{equation}
Unlike stationary kernels used in nonparametric regression (e.g., the Radial Basis Function; see \citealp{wand1994kernel}) that depend only on the Euclidean distance $\|x_i - x_j\|$, this kernel is non-stationary: it learns a metric in which similarity is determined by alignment of the projected features.

Under this definition, $A_{ij}$ is precisely the normalised kernel weight, and $Y_i$ is the Nadaraya–Watson estimator of the values $V$ at the query location $x_i$, conditioned on the context $\{x_j\}$:
\begin{equation}
 Y_i \;=\; \mathbb{E}_{\hat{P}}[V \mid x_i] \;=\; \sum_{j=1}^N \frac{\kappa_\eta(x_i, x_j)}{\sum_{\ell=1}^N \kappa_\eta(x_i, x_\ell)} v_j.
\end{equation}
Statistically, the self-attention layer performs in-context kernel smoothing: it integrates information across the value vectors $V$, with the smoothing bandwidth and shape dictated by the learned similarity structure $\kappa_\eta$.

This kernel-smoothing interpretation invites a comparison with Gaussian Process (GP) regression. For a GP with kernel $k(\cdot, \cdot)$ and noise variance $\sigma_n^2$, the posterior mean at a test point $x_*$ given training data $(X, \mathbf{y})$ is $m(x_*) = k(x_*, X) ( K(X, X) + \sigma_n^2 I )^{-1} \mathbf{y}$. Both architectures form a weighted sum of targets, but their mixing mechanisms differ. The GP posterior requires inverting the kernel matrix, an $\mathcal{O}(N^3)$ operation; attention replaces this inversion with a row-wise softmax that scales as $\mathcal{O}(N^2)$. Softmax normalization is only an approximation to the inversion step, but it allows the Transformer to scale to substantially longer sequences while retaining the capacity to model complex dependencies \citep{rasmussen2006gaussian}.

The query and key networks $\phi_Q, \phi_K$ also admit a spectral interpretation via Mercer's theorem. A Mercer kernel $\kappa(x, x^{\prime})$ admits the eigenfunction expansion $\sum_r \lambda_r \psi_r(x) \psi_r(x^{\prime})$, and the Transformer approximates this with a finite-rank inner product of feature maps,
\begin{equation}
\nonumber \phi_Q(x_i)\, \phi_K(x_j)^{\!\top} \;=\; \langle \phi_Q(x_i),\ \phi_K(x_j) \rangle,
\end{equation}
where $\phi_Q$ and $\phi_K$ are learned feature maps. Standard attention decouples these projections ($\phi_Q \neq \phi_K$)-- a choice motivated by causal or temporal asymmetry in language modeling--so the resulting attention matrix is not a Mercer kernel in the classical (symmetric) sense. Variants that tie the projections, such as the Reformer \citep{kitaev2020reformer}, recover symmetry by setting $\phi_Q = \phi_K$ and can then exploit locality-sensitive hashing \citep{andoni2008near}.

A full Transformer stack consists of multiple attention layers rather than a single one. This composition has a natural analog in Deep Gaussian Processes (DGPs; \citealp{damianou2013deep}), where the output of one Gaussian process serves as the input to the next, warping the data geometry layer by layer. Each attention layer computes a data-dependent kernel over the features extracted by the previous layer, so the depth of the stack enables the model to represent strongly non-stationary dependencies that a single shallow kernel cannot capture.

\section{Amortized Bayesian Inference}
\label{sec:amortized}

Amortized Bayesian inference is a computational strategy in which a substantial initial computational budget is allocated to training a global model, thereby making subsequent inference for individual test cases computationally efficient. This approach is particularly potent when combined with deep learning, replacing complex, iterative computational steps (such as MCMC or numerical optimization) with efficient forward passes of a neural network.

In many high-dimensional settings, such as image analysis or large-scale regression, the raw data $x$ often contains redundant information. Neural networks allow us to compress $x$ into lower-dimensional summary statistics, which classical statistical techniques can then leverage. In this work, we specifically focus on two complementary regimes of amortized inference: point estimation and uncertainty quantification.

\subsection{Point Estimation}
\label{subsec:point_est}

In the first regime, the neural network acts as a direct function approximator for a statistical estimator. The goal is to learn a mapping $f_\theta: \mathcal{X} \to \Theta$, parameterized by network weights $\theta$, that outputs a point estimate $\widehat{\vartheta}$ given observed data $x$. Throughout this section, we use $\vartheta \in \Theta$ for the inferential target (a generic Bayesian parameter) and reserve $\theta$ for the trainable weights of the amortized estimator, matching the convention in Section~\ref{sec:architecture}. This formulation reduces inference to a supervised regression problem.

To ensure the network approximates a valid Bayesian estimator, we employ a decision-theoretic framework. We define a loss function $L(\vartheta, \widehat{\vartheta})$ that quantifies the cost of estimating the true parameter $\vartheta$ as $\widehat{\vartheta}$, and minimize the Bayes risk--the expected loss over the joint distribution of parameters and data. Let $\pi(\vartheta)$ denote the prior over parameters and $p(x \mid \vartheta)$ the likelihood. The Bayes risk $\mathcal{R}(\pi, f_\theta)$ is
\begin{equation}
    \mathcal{R}(\pi, f_\theta) = \mathbb{E}_{\vartheta \sim \pi(\vartheta)} \left[ \mathbb{E}_{x \sim p(x \mid \vartheta)} \left[ L(\vartheta, f_\theta(x)) \right] \right],
\end{equation}
or equivalently
\begin{equation}
\nonumber    \mathcal{R}(\pi, f_\theta) = \int_\Theta \int_\mathcal{X} L(\vartheta, f_\theta(x))\, p(x \mid \vartheta)\, \pi(\vartheta) \, dx \, d\vartheta.
\end{equation}

Minimizing this risk with respect to $\theta$ yields the Bayes estimator under the chosen loss $L$. For instance, the squared-error loss $L(\vartheta, \widehat{\vartheta}) = \|\vartheta - \widehat{\vartheta}\|^2$ yields $f_{\theta^*}(x) \approx \mathbb{E}[\vartheta \mid x]$, the posterior mean.

\subsubsection{Implicit Topology Learning as Meta-Learning}

The capability of a neural network to perform statistical inference is not merely a curve-fitting exercise; it represents a form of meta-learning \citep{hospedales2022meta}. In traditional supervised learning, a model maps a specific input $x$ to a corresponding label $y$. In the amortized inference regime, the network learns a functional mapping from an entire dataset $\mathcal{D}_n = \{x_1, \dots, x_n\}$ to a parameter space $\Theta$.

This learning mechanism requires the network to internalize the topology of the parameter manifold. Rather than memorizing answers for specific datasets, the network must learn the inductive biases of the generative model--it learns \emph{to estimate}, not to recall. For example, when estimating the variance of a distribution, the network must implicitly learn that this parameter is strictly positive and that it correlates with the input dispersion, regardless of the specific input values.

Empirically, the latent representations learned by the network form a low-dimensional manifold aligned with the true parameter space $\Theta$. This geometric alignment allows the network to generalize to unseen datasets generated from the same prior, effectively replacing the manually derived formulas of classical estimation (such as the ordinary least squares formula) with a learned nonlinear equivalent.

\subsection{Uncertainty Quantification}
\label{subsec:uncertainty}

While point estimation provides a single optimal summary of the parameter space--typically corresponding to Maximum Likelihood Estimation (MLE) or Maximum A Posteriori (MAP) approximations \citep{bishop2006pattern, goodfellow2016deep}, it is inherently insufficient for capturing the full predictive or structural distribution. In complex modeling tasks, a single point estimate fails to capture the epistemic uncertainty arising from finite training data or the aleatoric uncertainty inherent in the data-generating process. Consequently, moving beyond deterministic predictions requires a coherent framework for full uncertainty quantification (UQ).

\paragraph{Bayesian Neural Networks:} 
Among the foundational paradigms for incorporating probabilistic reasoning into deep architectures is the Bayesian Neural Network (BNN) framework \citep{mackay1992practical, neal1996priors}. Instead of learning a single deterministic point estimate for the connection weights, a BNN assigns a prior probability distribution over the network parameters, denoted as $p(\mathbf{w})$. Given a training dataset $\mathcal{D}$, the inferential goal shifts to computing the posterior distribution over the weights, $p(\mathbf{w} \mid \mathcal{D})$, via Bayes' theorem. The predictive distribution for a novel test input $\mathbf{x}^*$ is subsequently evaluated by marginalizing over this parameter posterior:
$$p(y^* \mid \mathbf{x}^*, \mathcal{D}) = \int p(y^* \mid \mathbf{x}^*, \mathbf{w}) p(\mathbf{w} \mid \mathcal{D}) \, d\mathbf{w}$$
Despite their theoretical appeal, BNNs are computationally demanding and practically fragile. The true posterior over weights is analytically intractable because deep-network loss surfaces are high-dimensional and non-convex, so practitioners rely on structural or stochastic approximations. Sampling methods such as MCMC mix slowly in deep models and do not scale to architectures beyond shallow ones \citep{neal1996priors}. Optimization-based approaches such as Variational Inference \citep[VI;][]{graves2011practical, blundell2015weight} typically double the parameter count to model mean-field variances, introduce variance into gradient updates, and remain sensitive to hyperparameter tuning and initialization.

\paragraph{Deep Ensembles:} 
Another widely adopted paradigm for uncertainty quantification is the deep ensemble framework \citep{lakshminarayanan2017simple}. This approach involves training a collection of $M$ independent neural networks, with each model initialized from a distinct random seed and exposed to different data shuffles or splits. At deployment, the epistemic uncertainty is quantified by evaluating the variance across the predictions or estimated quantities of the individual ensemble members. From a theoretical standpoint, because the optimization landscape of a deep neural network is highly non-convex and high-dimensional, varying the initialization trajectory forces each network to settle into distinct local minima. Consequently, the distributed outputs generated by these independent architectures can be viewed as empirical samples drawn from an implicit, multimodal posterior distribution, effectively capturing the geometric landscape of the model's predictive uncertainty.

The principal bottleneck is optimization cost: training time and memory scale linearly with the ensemble size $M$, so deep ensembles are difficult to scale in large networks or resource-constrained deployment environments where repeatedly retraining the model is infeasible.

\paragraph{Amortized Variational Inference:} 
A third major paradigm for capturing uncertainty focuses on explicitly approximating the posterior distribution by restricting it to a tractable parametric family. Rather than executing independent optimization or sampling routines for each individual dataset, amortized variational inference (AVI) trains a global neural network to learn a direct functional mapping from the observed data to the parameters of a chosen variational distribution, $q_\phi(\boldsymbol{\theta} \mid \mathbf{x})$ \citep{kingma2013auto, gershman2014amortized}. During the upfront training phase, the network is optimized by maximizing the Evidence Lower Bound (ELBO) or minimizing an explicit probabilistic divergence (such as the Kullback-Leibler divergence) across a wide distribution of tasks. While traditional implementations often rely on simple mean-field Gaussian approximations, modern formulations extend this framework by integrating highly flexible density estimators, such as normalizing flows, to represent complex, skewed, or multimodal posterior surfaces \citep{rezende2015variational, papamakarios2021normalizing}.

A primary advantage of this methodology is its computational efficiency during deployment. It shifts the intensive inductive workload into a single offline training phase, spreading computational costs across all subsequent inference tasks. After optimizing the global network, evaluating the full approximate posterior or generating samples for a new dataset requires only one rapid forward pass through the network. This approach avoids the slow, iterative updates of classical MCMC methods and eliminates the large storage requirements of ensemble techniques.

\paragraph{Uncertainty Methodologies Utilized in this Study:}
To systematically evaluate and quantify uncertainty within our experimental analysis without incurring the computational overhead of deep ensembles or the optimization volatility of standard BNNs, we utilize three distinct methodologies: Monte Carlo Dropout, Monte Carlo Stability Analysis, and Flow-Based Posterior Sampling.

\paragraph{Monte Carlo Dropout:}
Dropout \citep{srivastava2014dropout} was introduced as a regularization technique used strictly during training: a random fraction of each layer's activations is zeroed at every step, which discourages co-adaptation of units and reduces overfitting. \citet{gal2016dropout} showed that retaining dropout at inference time yields a principled approximation to Bayesian inference. By keeping dropout active and repeating the forward pass on the same input, one obtains an empirical predictive distribution; the resulting stochastic forward passes are equivalent to variational inference in a Deep Gaussian Process, grounding a heuristic regularizer in formal inferential theory.

Formally, let $\widehat{\mathbf{W}}_l$ denote the trained weight matrix for layer $l \in \{1, \dots, L\}$. At inference time, for each stochastic forward pass $t \in \{1, \dots, T\}$, we sample a random binary mask vector from a Bernoulli distribution:
$$\mathbf{z}_{l, t} \sim \text{Bernoulli}(1-q)$$
where $q$ is the user-defined dropout probability. The active weights for the $t$-th realization are computed via the element-wise multiplication $\mathbf{W}_{l, t} = \widehat{\mathbf{W}}_l \cdot \text{diag}(\mathbf{z}_{l, t})$. Given a novel test instance $\mathbf{x}^*$, the predictive posterior distribution is empirically approximated by averaging across the collection of $T$ stochastic model trajectories:
$$p(y^* \mid \mathbf{x}^*, \mathcal{D}) \approx \frac{1}{T} \sum_{t=1}^T p(y^* \mid \mathbf{x}^*, \mathbf{W}_t)$$
From this sampled distribution, the epistemic uncertainty can be quantified by calculating the empirical variance across the $T$ predictions.

Despite its convenience, Monte Carlo (MC) dropout has well-known limitations. As a variational approximation, it inherits the standard failure modes of mean-field VI: a tendency to underestimate posterior variance and an inability to represent multimodal weight distributions, both of which are constrained by the rigid Bernoulli masking. MC dropout is also sensitive to model capacity: in small or shallow architectures, removing a fixed fraction of units degrades predictive capacity rather than calibrating uncertainty.

\paragraph{Monte Carlo Stability Analysis:}
\label{subsubsec:bootstrap}

To evaluate the consistency of our neural estimator, we employ a simulation-based stability analysis. We leverage the generative nature of our experimental framework to assess estimator variance across repeated realizations of the data-generation process.

For a fixed ground-truth $p$-length parameter vector $\beta_{\text{true}}$ and a specific sample size $N$, we generate $B$ independent datasets $\mathcal{D}_1, \dots, \mathcal{D}_B$, where each $\mathcal{D}_b = \{ (X^{(b)}, y^{(b)}) \}$ for $b=1,\ldots,B$. These datasets represent potential realizations of the world under identical underlying laws but different noise instances. We pass each dataset through the trained Transformer model to obtain a set of parameter estimates $\{\hat{\beta}^{(1)}, \dots, \hat{\beta}^{(B)}\}$. It is essential to note that our model employs a hard-thresholding mechanism to enforce sparsity; the network produces both a magnitude vector and an inclusion probability. The final estimate is computed as the element-wise product of the magnitude and a binary mask derived from the probabilities (thresholded at 0.5). We quantify uncertainty by computing the standard deviation of these estimates across the $B$ replications. The empirical stability metric is defined as the average standard deviation across all $p$ coefficients:
\begin{equation} \label{eq:bootmetric}
    \sigma_{\mathrm{MC}}(N) = \frac{1}{p} \sum_{j=1}^p \sqrt{ \frac{1}{B-1} \sum_{b=1}^B \left( \hat{\beta}_j^{(b)} - \bar{\beta}_j \right)^2 }.
\end{equation}
By evaluating the metric in \eqref{eq:bootmetric} across varying sample sizes (e.g., $N \in \{50, \dots, 1000\}$), we can empirically check the statistical consistency of the Transformer. A monotone decrease of $\sigma_{\mathrm{MC}}$ with $N$ is consistent with a $\sqrt{N}$-style information-aggregation rate.

\paragraph{Flow-Based Posterior Sampling:}
\label{subsubsec:flow_sampler}

While the Monte Carlo stability analysis quantifies the stability of the point estimate, it does not capture the whole geometry of the posterior distribution, particularly in multimodal settings. To address this, we employ a Flow Matching framework to train a neural network as a generative model that draws samples directly from the posterior $p(\theta \mid x)$. Further, to overcome the computational bottlenecks of traditional sampling, we leverage Conditional Flow Matching \citep[CFM,][]{lipman2023flow}. In contrast to MCMC samplers, which rely on local iterative updates to explore the parameter space, our flow-based mechanism learns a deterministic transformation from a simple base distribution to the complex target distribution.

The core premise is to identify a transformation capable of mapping samples from a tractable initial distribution $p_0$ (e.g., a standard Gaussian) to the intractable target posterior $p_1$. Flow matching models this transformation by identifying an ideal vector field that dictates the optimal movement of probability mass from the initial state to the final state over a time horizon $t \in [0, 1]$. We define a time-dependent vector field $v_t: \mathbb{R}^d \to \mathbb{R}^d$ parameterized by a neural network with parameters $\phi$. The evolution of a sample $z_t$ is governed by the ordinary differential equation
\begin{equation}
 \nonumber \frac{d z_t}{dt} = v_t(z_t; \phi).
\end{equation}
By integrating this derivative from $t=0$ to $t=1$, the neural network effectively transports samples from the prior directly to the posterior.

During training, we regress the neural vector field $v_t$ against a conditional target vector field $u_t$. We select the Optimal Transport path \citep{peyre2019computational}, which creates a straight-line trajectory between a noise sample $z_0$ and a data sample $z_1$. The loss function minimizes the discrepancy between the learned field and this ideal trajectory
\begin{equation}
    \mathcal{L}_{\mathrm{CFM}}(\phi) = \mathbb{E}_{t, z_0, z_1} \left[ \| v_t(z_t) - (z_1 - z_0) \|^2 \right].
\end{equation}
This approach effectively amortizes the inference cost: once the ideal vector field is learned, sampling becomes a fast, non-iterative forward pass through the ODE solver, avoiding the slow mixing and mode collapse issues often encountered with MCMC in complex topologies.

\section{Experimental Evaluation}
\label{sec:simulation}
\subsection{Experiment I: Latent Structure Recovery}
\label{exp:latent_structure}

We study task heterogeneity through a clustered task prior and evaluate whether permutation-invariant neural estimators can amortize inference by learning shared latent structure across tasks. Concretely, we construct a meta-learning environment where each task corresponds to a linear regression problem with a task-specific coefficient vector $\beta_t \in \mathbb{R}^p$, but the collection $\{\beta_t\}$ is constrained to lie on a finite set of latent centroids. We compare two amortized estimators, Deep Sets and Set Transformer, against a classical per-task estimator.

\paragraph{Generative process.}
We fix the feature dimension $p=20$ and the number of latent clusters $K$. We first sample $K$ latent centroids following $\mu_k \sim \mathcal{N}(0, \tau^2 I_p), k=1,\dots,K$ with $ \tau = 3.0$. These centroids remain fixed for the entire experiment and define a latent task mixture. For each task $t$, we sample its regression coefficient vector by selecting one centroid uniformly at random, i.e., $\beta_t \sim \text{Uniform}\left(\{\mu_1,\dots,\mu_K\}\right)$. Conditioned on $\beta_t$, we generate a task support set $\mathcal{D}_t = \{(x_{t,n},y_{t,n})\}_{n=1}^{N_t}$ with random support size $N_t \sim \mathrm{Discrete-Uniform}\{10,11,\dots,30\}$. We sample the inputs as $x_{t,n} \overset{\textrm{IID}}{\sim} \mathcal{N}(0,I_p)$ and outputs follow the linear model
\begin{equation}
\nonumber    y_{t,n} = x_{t,n}^\top \beta_t + \varepsilon_{t,n},
    \qquad \varepsilon_{t,n} \overset{\textrm{IID}}{\sim} \mathcal{N}(0,\sigma^2),
    \qquad \sigma = 1.
\end{equation}

\paragraph{Training and test meta-datasets.}
For each choice of $K\in\{5,10,50\}$, we independently generate a training meta-dataset of $T_{\mathrm{train}}$ tasks and a test meta-dataset of $T_{\mathrm{test}}=200$ held-out tasks using the same generative process (i.e., the same $K$ and hyperparameters). The neural models are trained across tasks to learn an amortized mapping
\begin{equation}
\nonumber    f_\theta:\ \mathcal{D}_t \mapsto \widehat{\beta}_t \in \mathbb{R}^p,
\end{equation}
where $f_\theta$ is expressed as either a Deep Sets encoder or a Set Transformer encoder followed by a regression head. Both architectures operate on the set of paired observations $\{(x_{t,n},y_{t,n})\}_{n=1}^{N_t}$ and are therefore permutation-invariant in the ordering of samples within a task.

\paragraph{Baselines and evaluation metric.}
As classical baselines, we compute a per-task Ordinary Least Squares (OLS) estimate $\widehat{\beta}^{\mathrm{OLS}}_t$ and a Ridge regression estimate $\widehat{\beta}^{\mathrm{Ridge}}_t$ using only the support set $\mathcal{D}_t$. We evaluate all estimators on the test tasks using parameter estimation error
\begin{equation} \label{eq:mse_beta}
    \mathrm{MSE}_\beta \;=\; \frac{1}{T_{\mathrm{test}}}\sum_{t=1}^{T_{\mathrm{test}}}\left\|\widehat{\beta}_t - \beta_t\right\|_2^2,
\end{equation}
and train neural models for 100 epochs by minimizing the same objective over training tasks.

\paragraph{Parameter estimation error.}
Table~\ref{tab:results_mse} reports parameter-estimation error in the latent-structure-recovery experiment. The amortized neural estimators achieve parameter MSE one to two orders of magnitude below the per-task Ridge baseline. Linear baselines (OLS and Ridge) cannot exploit the shared geometry across tasks, so they suffer high error regardless of the cluster count $K$. Ridge improves marginally over OLS through $\ell_2$ regularization, but remains fundamentally per-task. In low-complexity regimes ($K=5, 10$), the neural estimators closely approximate the Bayes-optimal estimator; at $K=50$, both architectures continue to generalize, indicating that they exploit the finite cluster structure of the parameter space even as it becomes more diffuse.

\begin{table}[t]
\centering
\caption{Parameter estimation error ($\mathrm{MSE}_\beta$ in \eqref{eq:mse_beta}), standard error (SE), and 95\% coverage (Cov95) in the latent structure recovery experiment. Results are averaged over 200 held-out test tasks.
For Ridge, we consider per-task linear estimators trained independently on each task using only its support set. For Deep Sets and Set Transformer, we consider amortized estimators trained across tasks. \vspace{2mm}
}
\label{tab:results_mse}
\begin{tabular}{lccccc}
\toprule
& Ridge (per-task) & \multicolumn{2}{c}{Deep Sets} & \multicolumn{2}{c}{Set Transformer} \\
\cmidrule(lr){2-2} \cmidrule(lr){3-4} \cmidrule(lr){5-6}
Latent complexity & MSE & MSE $\pm$ SE & Cov95 & MSE $\pm$ SE & Cov95 \\
\midrule
Clusters ($K=5$)  & $5.027$ & $0.248 \pm 0.016$ & $0.980$ & $\mathbf{0.043} \pm 0.002$ & $0.999$ \\
Clusters ($K=10$) & $5.245$ & $1.031 \pm 0.047$ & $0.928$ & $\mathbf{0.102} \pm 0.007$ & $0.980$ \\
Clusters ($K=50$) & $4.822$ & $6.242 \pm 0.169$ & $0.315$ & $\mathbf{2.370} \pm 0.100$ & $0.623$ \\
\bottomrule
\end{tabular}
\end{table}

\subsection{Experiment II: Distributional Robustness}
\label{exp:asymptotic_robustness}

We next study the behavior of amortized estimators as the meta-training-set size grows, and their robustness to distributional mismatch. This experiment jointly evaluates (i) sample complexity in the number of training tasks and (ii) stability under increasingly non-Gaussian noise distributions. Unlike Experiment I, where task heterogeneity is induced through a clustered prior, here tasks are independently generated but exhibit substantial within-task uncertainty due to high-variance parameters and complex residual structure.

\paragraph{Generative process.}
We fix the feature dimension $p=20$. For each task $t$, we draw regression coefficients independently from a high-variance prior $\beta_t \overset{\textrm{IID}}{\sim} \mathcal{N}(0, 9^2 I_p)$. Conditioned on $\beta_t$, we generate a support set $\mathcal{D}_t = \{(x_{t,n},y_{t,n})\}_{n=1}^{N_t}$ with inputs sampled as $x_{t,n} \overset{\textrm{IID}}{\sim} \mathcal{N}(0, I_p)$. We generate outputs according to the linear model
$$y_{t,n} = x_{t,n}^\top \beta_t + \varepsilon_{t,n},$$
where the noise distribution $\varepsilon_{t,n}$ varies across regimes described below. To rigorously evaluate robustness to distributional mismatch, the final evaluation is performed on a held-out test set of $T_{\mathrm{test}} = 600$ independent tasks for each noise regime.

\paragraph{Noise regimes.}
To assess robustness to distributional mismatch, we consider four residual distributions: (I) Gaussian (symmetric): $\varepsilon_{t,n} \sim \mathcal{N}(0,1)$, (II) Asymmetric: $\varepsilon_{t,n} \overset{d}{=} E - 1$, where $E \sim \mathrm{Exp}(1)$, (III) Bimodal: $\varepsilon_{t,n}$ follows a two-component Gaussian mixture with weights $\{0.8, 0.2\}$, and (IV) Trimodal: $\varepsilon_{t,n}$ follows a three-component Gaussian mixture with weights $\{0.8, 0.1, 0.1\}$.
For the multimodal cases, component-specific means and variances are chosen to ensure well-separated modes, inducing heavy-tailed and non-convex error distributions.

\paragraph{Training and evaluation protocol.}
We generate a meta-dataset of up to $T=6000$ tasks using the generative process above. From this pool, we form meta-training sets of size $T_{\mathrm{train}} \in \{5,\,10,\,30,\,50,\,80,\,100,\,300,\,400\}$ and a held-out test set of $T_{\mathrm{test}}=600$ tasks. For each value of $T_{\mathrm{train}}$, we train Deep Sets and the Set Transformer to convergence using the same fixed optimization budget per fit, and evaluate the resulting estimator on the held-out test tasks.

This protocol examines sample complexity as a function of the meta-training-set size. Robustness to distributional mismatch is assessed by evaluating each trained model under every residual regime (Gaussian, Asymmetric, Bimodal, and Trimodal) while keeping the task prior and the task sample-size distribution fixed.

\paragraph{Objective and metrics.}
As in Experiment I, Deep Sets and Set Transformer architectures learn an amortized mapping $f_\theta:\ \mathcal{D}_t \mapsto \widehat{\beta}_t \in \mathbb{R}^p$. We evaluate the performance using the parameter estimation error $\mathrm{MSE}_\beta = \mathbb{E}\left[\|\widehat{\beta}_t - \beta_t\|_2^2\right]$, computed over held-out tasks and averaged across noise regimes and support sizes.

\begin{figure}[t]
    \centering
    \begin{subfigure}[t]{0.48\textwidth}
        \centering
        \includegraphics[width=\linewidth]{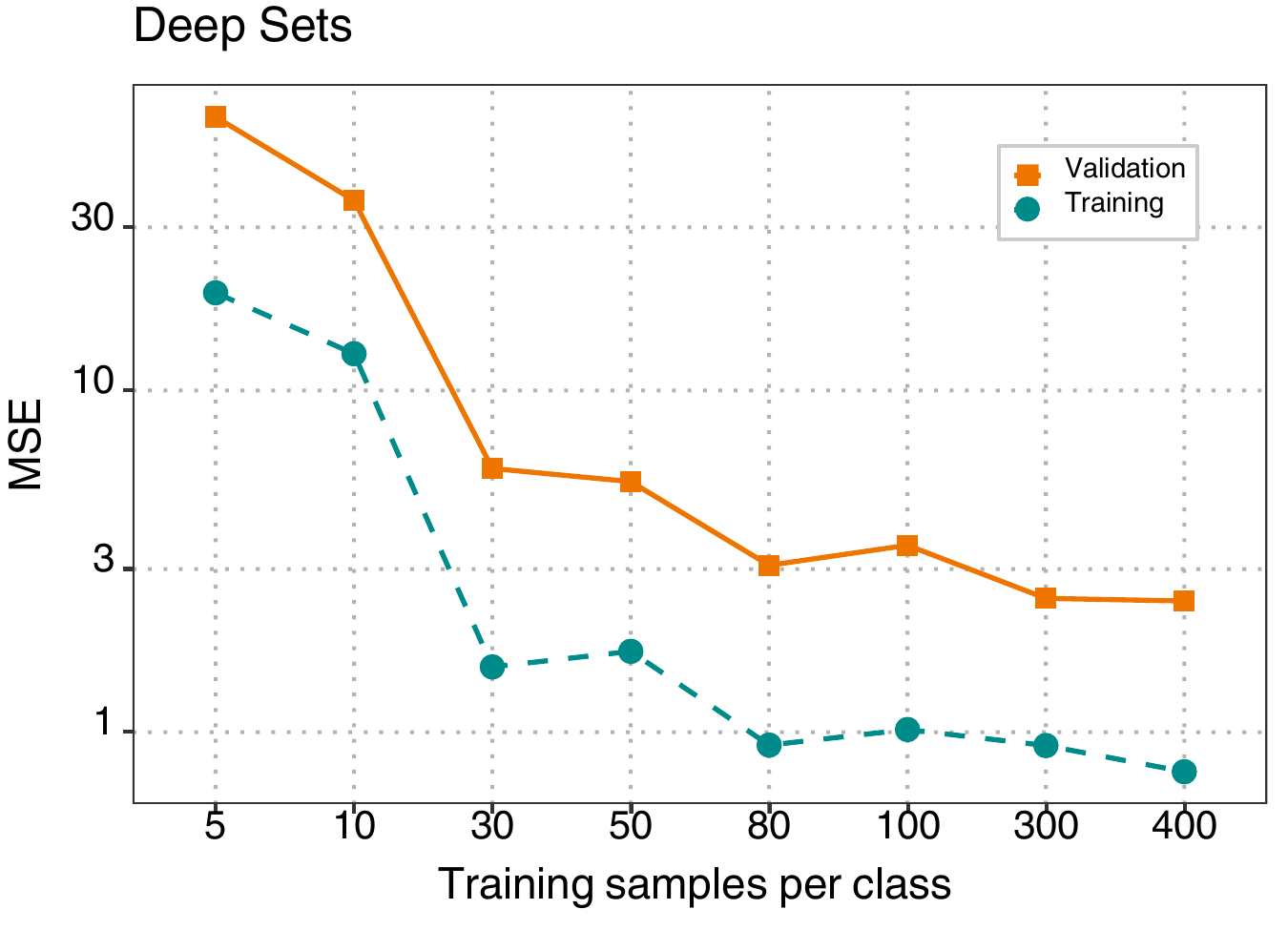}
        \caption{Deep Sets}
        \label{fig:deepsets_dynamics}
    \end{subfigure}
    \hfill
    \begin{subfigure}[t]{0.48\textwidth}
        \centering
        \includegraphics[width=\linewidth]{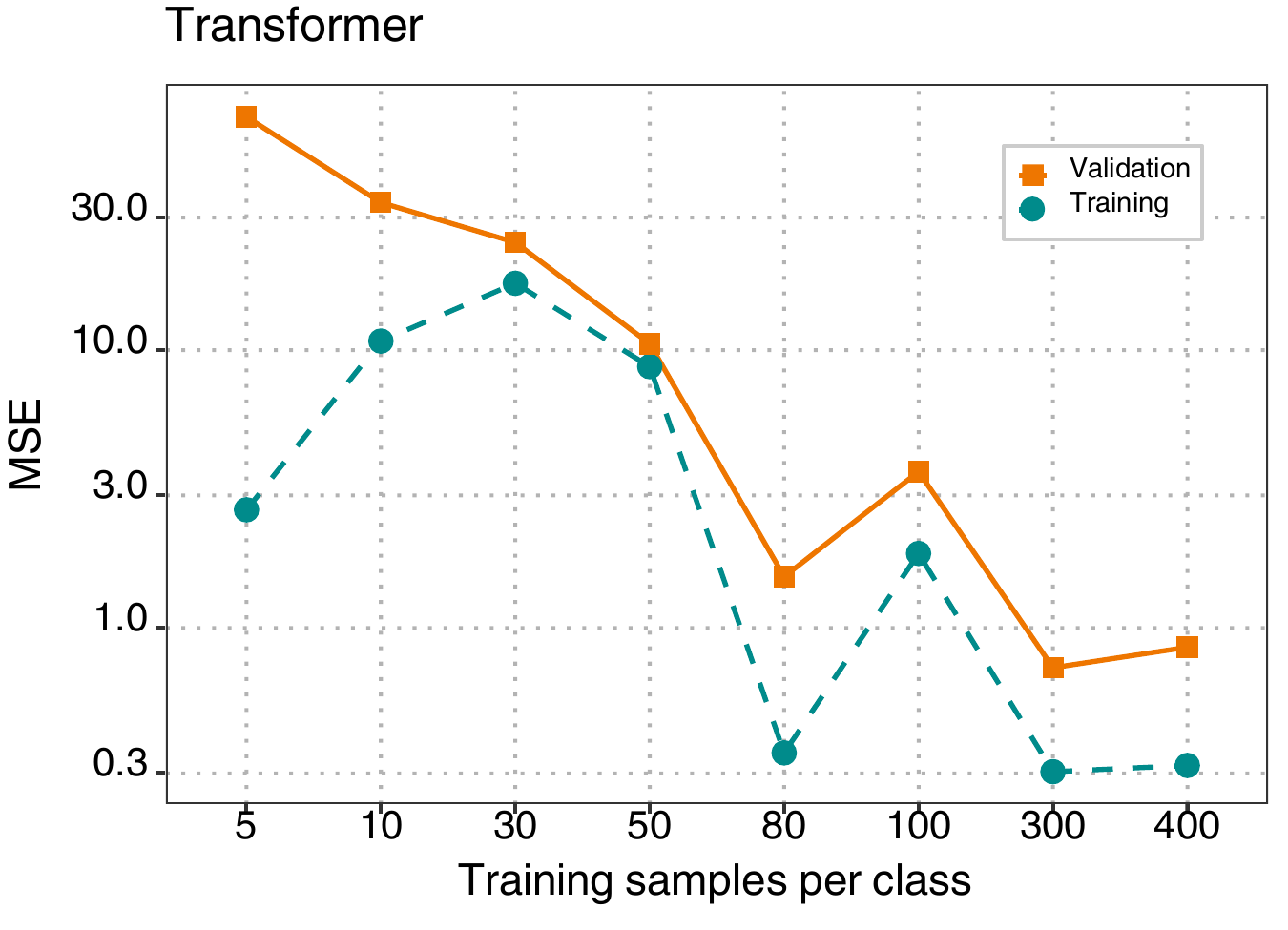}
        \caption{Set Transformer}
        \label{fig:transformer_dynamics}
    \end{subfigure}
    \caption{Sample complexity in the meta-training-set size: dashed lines report training-task fit and solid lines report held-out test error, as a function of the number of training tasks $T_{\mathrm{train}}$. (\subref{fig:deepsets_dynamics}) Deep Sets achieves low error even with small $T_{\mathrm{train}} < 50$ but plateaus quickly. (\subref{fig:transformer_dynamics}) The Set Transformer requires more training tasks ($T_{\mathrm{train}} > 80$) to stabilize, but reaches a substantially lower asymptotic error.}
    \label{fig:learning_dynamics}
\end{figure}

\paragraph{Sample complexity in the meta-training-set size.}
Figure~\ref{fig:learning_dynamics} reports the training-task fit and held-out test error of Deep Sets and the Set Transformer as the meta-training-set size $T_{\mathrm{train}}$ is varied over $\{5, 10, 30, 50, 80, 100, 300, 400\}$. The horizontal axis is the number of training tasks on which the amortized estimator was fit. It is distinct from the optimization budget (held fixed per fit) and from the per-task sample size $N_t$ (drawn from a fixed distribution).

Deep Sets (left panel) attains low test error with only a small number of training tasks; both training and held-out error decrease sharply for $T_{\mathrm{train}} \le 30$--$50$ and improve only marginally thereafter. The Set Transformer (right panel) has higher error and greater variance in the small-$T_{\mathrm{train}}$ regime, but its error continues to decrease as more training tasks are supplied. By $T_{\mathrm{train}} \ge 300$, the Set Transformer reaches a substantially lower asymptotic error than Deep Sets. In summary, Deep Sets is sample-efficient in $T_{\mathrm{train}}$ but capacity-limited; the Set Transformer requires more training tasks to identify but attains a better solution.
\paragraph{Complex noise robustness.} 
To systematically evaluate the robustness of the models, we injected different noise profiles into the dataset. The noise distributions in Table \ref{tab:noise_results} are categorized into three distinct groups: \textit{Clean} (symmetric, baseline Gaussian noise), \textit{Skewed} (asymmetric, heavy-tailed noise that violates standard normality assumptions), and \textit{Multimodal} (mixture models introducing distinct, unequally weighted error clusters).

\begin{table}[t]
\centering
\caption{Parameter estimation error ($\mathrm{MSE} \pm \mathrm{SE}$) and 95\% coverage across clean, skewed, and multimodal noise distributions. \textit{Clean Small} and \textit{Clean High} denote Gaussian residuals with low and high variance ($\sigma=1$ and $\sigma=3$). \textit{Right/Left Skew (r1)} and \textit{(r2)} are shifted Exponential residuals with rate parameters $r_1=1$ and $r_2=0.5$ (heavier-tailed). \textit{Bi-Unbalanced (m$\mu$s$\sigma$)} denotes a two-component Gaussian mixture with weights $\{0.8, 0.2\}$ whose minority component has mean $\mu$ and standard deviation $\sigma$. \textit{Trimodal Tight} and \textit{Trimodal Wide} denote three-component Gaussian mixtures with weights $\{0.8, 0.1, 0.1\}$ whose modes are closely or widely separated, respectively.}
\label{tab:noise_results}
\begin{tabular}{lcccc}
\toprule
& \multicolumn{2}{c}{Deep Sets} & \multicolumn{2}{c}{Set Transformer} \\
\cmidrule(lr){2-3} \cmidrule(lr){4-5}
Noise Type & MSE $\pm$ SE & Cov95 & MSE $\pm$ SE & Cov95 \\
\midrule
\multicolumn{5}{l}{\textit{Clean Distributions}} \\
Clean Small & $6.86 \pm 0.357$ & $0.980$ & $1.72 \pm 0.100$ & $0.984$ \\
Clean High  & $7.17 \pm 0.341$ & $0.974$ & $1.56 \pm 0.081$ & $0.991$ \\
\midrule
\multicolumn{5}{l}{\textit{Skewed Distributions}} \\
Right Skew (r1) & $7.61 \pm 0.393$ & $0.972$ & $1.65 \pm 0.092$ & $0.988$ \\
Left Skew (r1)  & $7.20 \pm 0.340$ & $0.975$ & $1.63 \pm 0.100$ & $0.989$ \\
Right Skew (r2) & $6.78 \pm 0.339$ & $0.977$ & $1.54 \pm 0.082$ & $0.989$ \\
Left Skew (r2)  & $8.00 \pm 0.404$ & $0.974$ & $1.73 \pm 0.108$ & $0.986$ \\
\midrule
\multicolumn{5}{l}{\textit{Multimodal Distributions}} \\
Bi-Unbalanced (m3s1) & $7.96 \pm 0.414$ & $0.972$ & $2.21 \pm 0.125$ & $0.977$ \\
Bi-Unbalanced (m4s2) & $9.99 \pm 0.451$ & $0.953$ & $3.36 \pm 0.201$ & $0.955$ \\
Trimodal Tight       & $8.00 \pm 0.377$ & $0.975$ & $1.95 \pm 0.112$ & $0.985$ \\
Trimodal Wide        & $7.42 \pm 0.385$ & $0.970$ & $1.72 \pm 0.089$ & $0.990$ \\
\bottomrule
\end{tabular}
\end{table}

Training the Set Transformer is more demanding than training simpler architectures: it requires a larger meta-training set and careful hyperparameter tuning, and we observed slower and less stable convergence under heavy-tailed and multimodal noise. The high coverage reported for the Set Transformer in Table~\ref{tab:noise_results} is partly explained by the wider predictive variance due to this weight sensitivity, as reflected in the larger standard errors of its MSE estimates.

\begin{figure}[t]
    \centering
    \includegraphics[width=0.95\textwidth]{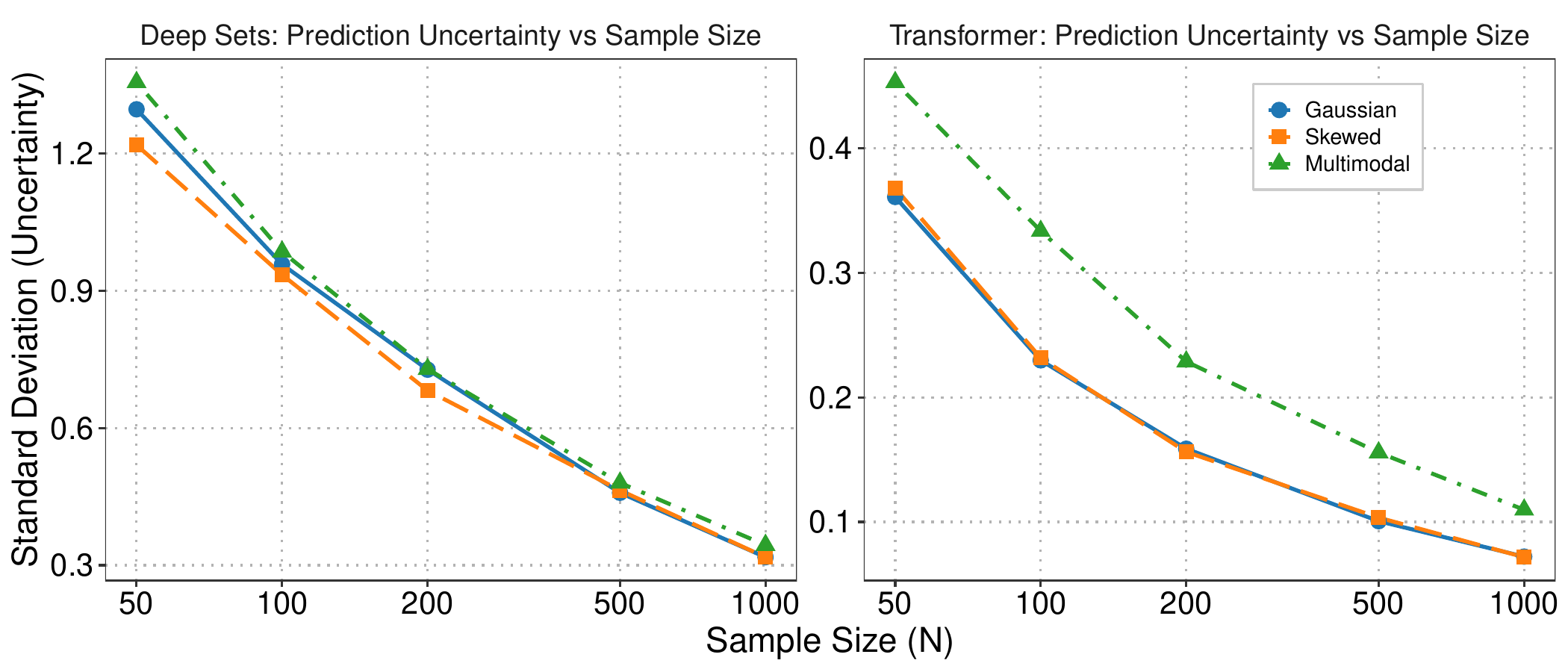}
    \vspace{-3mm}
    \caption{
    Monte Carlo stability analysis. Standard deviation of the estimated regression coefficients (computed via replicated sampling of the task support set) as a function of support set size $N$. Both architectures exhibit decreasing uncertainty as $N$ increases.
    }
    \label{fig:uncertainty_plots}
\end{figure}

As shown in Figure~\ref{fig:uncertainty_plots}, we assess the stability of the amortized estimators by measuring Monte Carlo variability as a function of the support set size $N$. Here, $N$ denotes the number of observations provided to the model for a single task at inference time, and should be clearly distinguished from the meta-training-set size $T_{\mathrm{train}}$ or the optimization budget. For each task and noise regime, we draw replicated samples over the support set and compute the standard deviation of the resulting parameter estimates. This quantity captures the sensitivity of the estimator to sampling variability and provides a direct measure of predictive stability. Both architectures exhibit a clear monotonic decrease in Monte Carlo standard deviation as $N$ increases, indicating that parameter estimates become more stable as additional evidence is provided. This behavior is consistent with that of well-behaved statistical estimators, where the uncertainty decreases as the information content of the observed data increases. Notably, the Set Transformer demonstrates substantially lower uncertainty across all noise regimes and support sizes. In particular, its Monte Carlo standard deviation remains below $0.6$ even in low-data settings, whereas Deep Sets exhibits considerably higher variability, with deviations exceeding $1.5$ for small $N$. Moreover, the Transformer's uncertainty curves for Gaussian, asymmetric, and multimodal noise remain closely aligned, suggesting that its stability is largely insensitive to the form of the noise distribution. This observation indicates that the attention-based architecture produces more reliable and consistent parameter estimates under both data scarcity and distributional mismatch.

\subsection{Experiment III: Sparse Signal Recovery}
\label{exp:sparse_recovery}

To further probe Regime II (the data-scarce frontier), we study the recovery of sparse signals in high-dimensional linear regression. Each task corresponds to estimating a task-specific sparse coefficient vector from a small set of observations. Our goal is to learn an amortized, permutation-invariant estimator that identifies both the support and magnitudes of the sparse signal.

\paragraph{Generative process.}
We fix the feature dimension $p=100$. For each task $t$, we generate a sparse regression coefficient vector $\beta_t\in\mathbb{R}^p$ with a sparsity level $k_t$ denoting the \emph{percentage of zero elements}, so that
\[
\|\beta_t\|_0 = p - k_t,
\qquad k_t \in \mathcal{K} := \{5,10,15,\dots,100\}.
\]
The active support $S_t\subset\{1,\dots,p\}$ with $|S_t|=p-k_t$ is sampled uniformly at random (e.g., $k_t = 95$ leaves only $5$ active variables). Conditional on $S_t$, the nonzero coefficients are drawn independently as
$(\beta_t)_j \sim \mathcal{N}(0,\,3), \quad j\in S_t$, with $(\beta_t)_j=0$ for $j\notin S_t$. Conditioned on $\beta_t$, we generate a task-specific support set $\mathcal{D}_t=\{(x_{t,n},y_{t,n})\}_{n=1}^{N_t}$, where covariates are sampled as $x_{t,n}\sim \mathcal{N}(0,I_p)$ and responses follow the linear model
\[
y_{t,n}=x_{t,n}^\top \beta_t+\varepsilon_{t,n},
\qquad \varepsilon_{t,n}\sim \mathcal{N}(0,1).
\]
The number of observations per task is drawn uniformly as
\[
N_t \sim \mathrm{Discrete-Uniform}\{400,401,\dots,500\}.
\]

\paragraph{Model.}
For each task $t$, we treat $\mathcal{D}_t$ as an unordered set of tokens $z_{t,n}=[x_{t,n};y_{t,n}] \in \mathbb{R}^{p+1}$. Tokens are embedded into $\mathbb{R}^{d_{\text{model}}}$ and processed by an $L$-layer Transformer encoder `without positional encodings', ensuring permutation equivariance across samples. A mean pooling operator produces a permutation-invariant task representation $r_t$, which is decoded by two heads: a magnitude head $\widehat{\beta}_{\mathrm{mag},t}\in\mathbb{R}^p$ and a sparsity gate $\hat p_t\in(0,1)^p$. The final coefficient estimate is
\[
\widehat{\beta}_t=\widehat{\beta}_{\mathrm{mag},t}\odot \hat p_t.
\]

\paragraph{Training objective.}
Given a batch of $M$ tasks, we minimize prediction error using soft gating:
\begin{equation}
\nonumber    \mathcal{L}  = \frac{1}{M} \sum_{t=1}^{M}
    \left(
        \frac{1}{N_t} \sum_{n=1}^{N_t}
        \left(
            y_{t,n} - x_{t,n}^\top \widehat{\beta}_t
        \right)^2
    \right).
\end{equation}

\paragraph{Training meta-dataset construction.}
We generate a fixed meta-dataset of $T=6000$ tasks spanning the grid of sparsity levels $\mathcal{K} = \{5,10,15,\dots,100\}$ ($|\mathcal{K}|=20$). For each $k\in\mathcal{K}$, we generate $300$ independent tasks, yielding $T = 20 \times 300 = 6000$ tasks in total. Each task $t$ is assigned a sparsity level $k_t \in \mathcal{K}$ and is generated according to the sparse linear model described above.

\paragraph{Per-task sampling.}
For a task with sparsity level $k_t$ (percentage of zero coefficients), we sample an active support $S_t\subset\{1,\dots,p\}$ with $|S_t|=p-k_t$ uniformly at random, draw nonzero coefficients $(\beta_t)_j\sim\mathcal{N}(0,1)$ for $j\in S_t$ (and $(\beta_t)_j=0$ otherwise), and then generate a task-specific dataset $\mathcal{D}_t=\{(x_{t,n},y_{t,n})\}_{n=1}^{N_t}$. The number of observations per task is drawn uniformly as $N_t \sim \mathrm{Discrete-Uniform}\{400,401,\dots,500\}$, with covariates sampled as $x_{t,n} \overset{\mathrm{IID}}{\sim} \mathcal{N}(0,I_p)$ and responses generated as
\[
y_{t,n}=x_{t,n}^\top\beta_t+\varepsilon_{t,n}, \qquad \varepsilon_{t,n}\sim\mathcal{N}(0,1).
\]

\paragraph{Train--test split.}
After generating the full collection of $T=6000$ tasks, we randomly shuffle task indices and split tasks into $T_{\mathrm{train}}=5400$ training tasks and $T_{\mathrm{test}}=600$ held-out test tasks. All models are trained only on the training tasks; evaluation is performed on the held-out tasks.

\begin{figure}[t]
    \centering
    \includegraphics[width=0.95\textwidth]{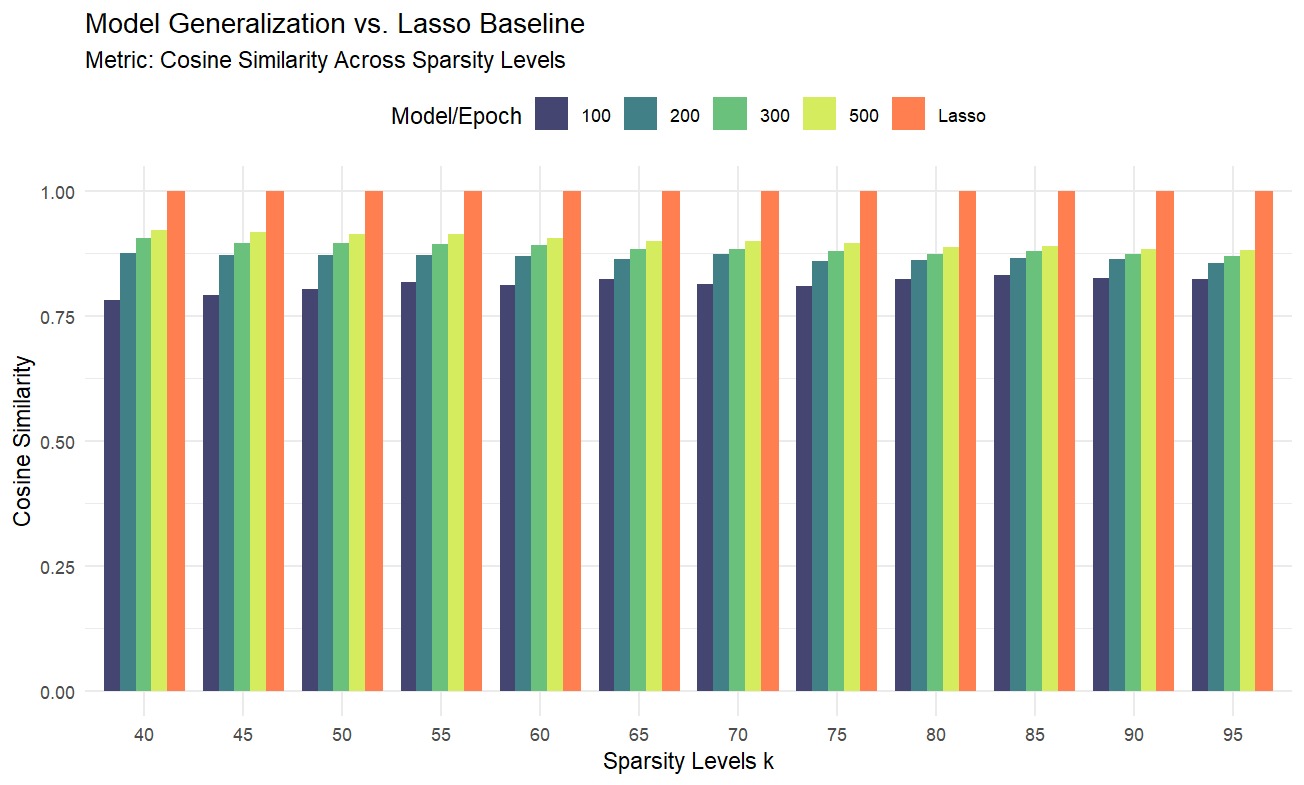} 
    \vspace{-3mm}
    \caption{
    Cosine similarity between estimated and true regression coefficients across different sparsity levels $k$, where $k\%$ of the coefficients in the ground-truth parameter vector are zero. Bar clusters compare the neural network's optimization progression at training epochs 100, 200, 300, and 500 against the baseline Lasso regression performance.
    }
    \label{fig:cosine_similarity}
\end{figure}
\paragraph{Performance.} Figure~\ref{fig:cosine_similarity} reports the across-epoch performance of the sparse recovery architecture under varying sparsity levels compared against the Lasso baseline. We measure cosine similarity between the predicted coefficient vector $\widehat{\boldsymbol{\beta}}$ and the ground-truth $\boldsymbol{\beta}$, where the sparsity parameter $k$ denotes the percentage of zero coefficients in $\boldsymbol{\beta}$. The Lasso baseline demonstrates exceptional point-estimation accuracy, consistently achieving a cosine similarity close to 1 across all sparsity regimes. For our neural model, moderate sparsity levels (e.g., $k \leq 60\%$) yield high cosine similarity ($>0.85$) even at early training stages (epoch 100), indicating accurate recovery of the dominant signal components. As sparsity increases (larger $k$), the recovery task becomes more challenging due to fewer informative features, leading to lower initial performance at epoch 100. However, continued optimization leads to consistent improvements, and by epoch 500, the model closely approaches the baseline performance. Crucially, while Lasso solves an independent optimization problem for each task at inference time, the amortized model produces estimates in a single forward pass, making it well-suited to large-scale multi-task deployment.

While cosine similarity captures point-estimation accuracy, it does not reflect the stability of the recovered coefficients across different data realizations. To assess the reliability of the estimator beyond point estimates, we analyze the empirical standard deviation of the predicted coefficients, denoted as $\sigma(\widehat{\boldsymbol{\beta}})$. Table~\ref{tab:std_sparsity_samplesize} reports this quantity as a function of the support set size $N$ across sparsity percentages $k \in \{20, 50, 80, 95\}$. In low-sample regimes ($N=50$), uncertainty is elevated across all sparsity levels, particularly when sparsity is high, reflecting the difficulty of disentangling signal from noise with limited observations. As the number of observations increases, the standard deviation decreases monotonically for all sparsity levels. At $N=1000$, uncertainty is uniformly low, indicating stable coefficient recovery even under extreme sparsity.

\begin{table}[t]
    \centering
    \caption{
    Standard deviation of predicted regression coefficients $\sigma(\widehat{\boldsymbol{\beta}})$ across varying support set sizes ($N$) and sparsity percentages ($k$).
    Here, $k$ denotes the percentage of zero coefficients in the true parameter vector. The values reported here are calculated based on averaging the standard deviations obtained from 30 replications.
    }
    \label{tab:std_sparsity_samplesize}
    \begin{tabular}{c c c c c}
    \toprule
    \multirow{2}{*}{Support set size ($N$)} & \multicolumn{4}{c}{Sparsity percentage ($k$)} \\
    \cmidrule(lr){2-5}
     & 20 & 50 & 80 & 95 \\
    \midrule
    50   & 0.530 & 0.498 & 0.445 & 0.412 \\
    100  & 0.392 & 0.355 & 0.310 & 0.285 \\
    200  & 0.258 & 0.220 & 0.185 & 0.150 \\
    500  & 0.140 & 0.125 & 0.095 & 0.085 \\
    1000 & 0.062 & 0.055 & 0.048 & 0.042 \\
    \bottomrule
\end{tabular}
\end{table}

\subsection{Experiment IV: Topological Mismatch in Multimodality}
\label{exp:ring_modes}

We next consider a setting in which the posterior distribution over task parameters exhibits a substantial topological mismatch relative to standard unimodal assumptions. In particular, we study amortized inference when the parameter distribution is highly multimodal and supported on a non-convex manifold. This experiment evaluates whether neural inference models can identify and represent disconnected posterior geometry using only observed input--output samples.

\paragraph{Generative model.}
Let $\boldsymbol{\beta} \in \mathbb{R}^2$ denote the task-specific regression parameter. We define a multimodal prior distribution consisting of $K=8$ Gaussian components arranged uniformly on a circle of radius $R=5.0$:
\begin{equation}
\nonumber    p(\boldsymbol{\beta})
    =
    \frac{1}{K}
    \sum_{k=1}^{K}
    \mathcal{N}(\boldsymbol{\beta} \mid \boldsymbol{\mu}_k, \sigma^2 I_2),
    \qquad
    \boldsymbol{\mu}_k
    =
    R
    \begin{bmatrix}
        \cos\left(\tfrac{2\pi k}{K}\right) \\
        \sin\left(\tfrac{2\pi k}{K}\right)
    \end{bmatrix}.
\end{equation}
Conditioned on $\boldsymbol{\beta}$, observations are generated according to the same linear model used throughout the paper
\begin{equation}
\nonumber    y_n = x_n^\top \boldsymbol{\beta} + \varepsilon_n,  \qquad
    x_n \sim \mathcal{N}(0, I_2), \;
    \varepsilon_n \sim \mathcal{N}(0, \sigma_\varepsilon^2),
\end{equation}
yielding a task-specific dataset $\mathcal{D} = \{(x_n, y_n)\}_{n=1}^{N}$.

\paragraph{Inference objective.}
Given $\mathcal{D}$, the goal is to approximate the posterior distribution $p(\boldsymbol{\beta} \mid \mathcal{D}),$
which inherits the multimodal structure of the prior while being shaped by the likelihood induced by the observed $(x,y)$ pairs. Notably, the inference model has access only to the observed samples and must identify the full posterior geometry without explicit knowledge of the prior components.

\paragraph{Amortized conditional flow.}
We employ a conditional normalizing flow trained via Conditional Flow Matching to learn a mapping from observed datasets to posterior distributions. The input dataset $\mathcal{D}$ is first embedded into a fixed-dimensional context vector
\[
\bm{r} = f_{\psi}(\mathcal{D}),
\]
using a permutation-invariant encoder consistent with earlier experiments (Deep Sets). This representation summarizes the information contained in the observed $(x,y)$ samples and conditions the inference model.

Let $\boldsymbol{\beta}_t \in \mathbb{R}^2$ denote the state of the flow at time $t \in [0,1]$, initialized from a base distribution $\boldsymbol{\beta}_0 \sim \mathcal{N}(0, I_2)$. The evolution of $\boldsymbol{\beta}_t$ is governed by a time-dependent velocity field
\begin{equation}
    \frac{d \boldsymbol{\beta}_t}{dt}
    =
    v_{\phi}(t, \boldsymbol{\beta}_t, \bm{r}),
\end{equation}
where $v_{\phi}$ is a neural network parameterized by $\phi$. Integrating this ordinary differential equation yields
\begin{equation}
    \boldsymbol{\beta}_1
    =
    \boldsymbol{\beta}_0
    +
    \int_{0}^{1}
    v_{\phi}(t, \boldsymbol{\beta}_t, \bm{r}) \, dt,
\end{equation}
which represents a sample from the learned approximation to $p(\boldsymbol{\beta} \mid \mathcal{D})$.

\paragraph{Posterior geometry and transport dynamics.}
\begin{figure}[t]
    \centering
    \includegraphics[width=\textwidth]{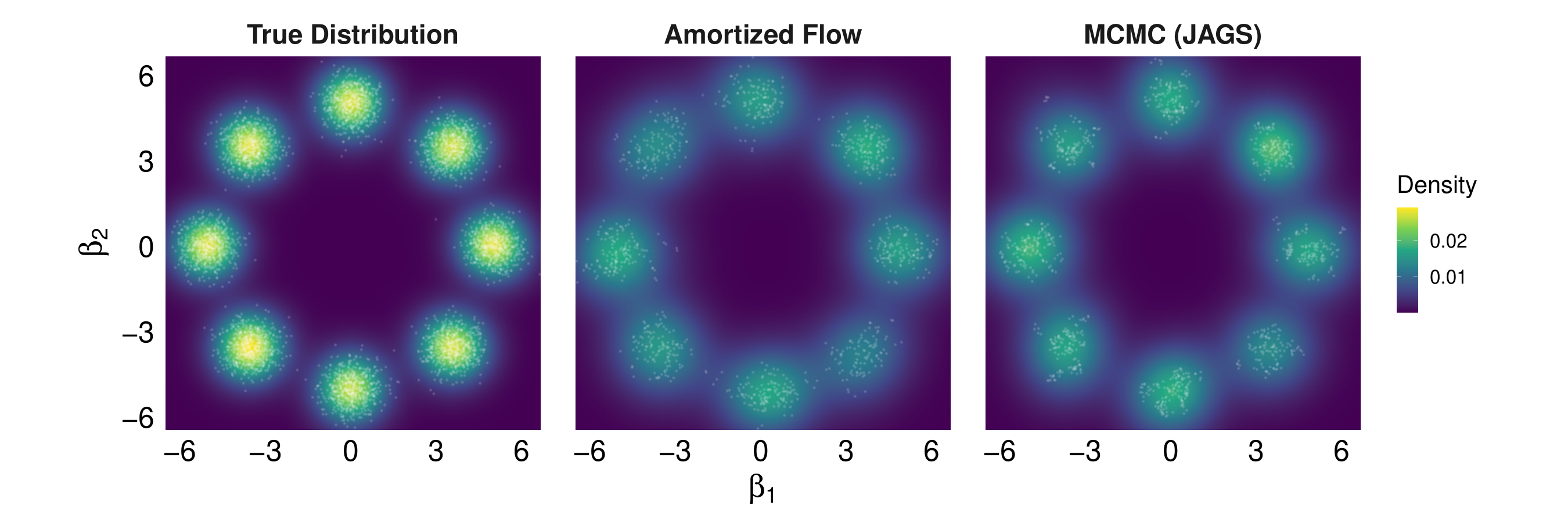}
    \caption{Posterior Sampling Benchmark: Visual comparison of the true multimodal distribution (Left), the Amortized Normalizing Flow approximation (Center), and the MCMC baseline (Right). The Amortized Flow successfully captures the complex multimodal geometry and separates the modes significantly faster than the baseline.}
    \label{fig:benchmark_comparison}
\end{figure}
Figure~\ref{fig:benchmark_comparison} compares samples from the true posterior distribution, the amortized flow approximation, and an MCMC baseline. The amortized model successfully allocates probability mass across all eight modes, demonstrating that it recovers the disconnected geometry of the posterior using only the observed $(x,y)$ samples. To quantitatively compare the posterior approximations, we compute the Kullback–Leibler (KL) and Jensen–Shannon (JS) divergences between the true posterior and each approximation. The amortized flow yields a KL divergence of 0.2728 and a JS divergence of 0.0587, while the corresponding values for the MCMC approximation are 0.2544 and 0.0528, respectively.

\begin{figure}[t]
    \centering    
    \includegraphics[width=\textwidth]{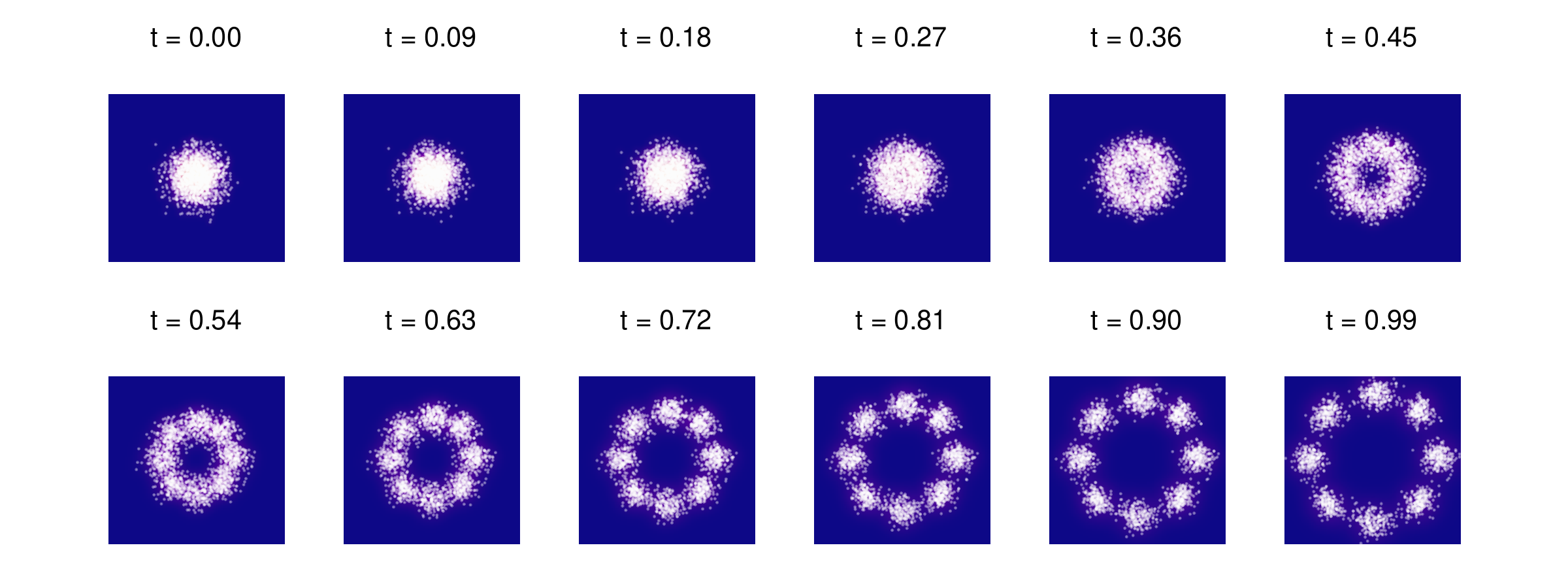}
    \caption{Visualization of the flow trajectory evolution. The horizontal panels illustrate the transformation of the initial particle distribution ($\theta_0 \sim \mathcal{N}(0, I)$ at $t=0.00$) under the learned vector field over time. By $t=1.00$, the particles have successfully converged to the target 8-mode posterior geometry.}
    \label{fig:flow_trajectory}
\end{figure}
Figure~\ref{fig:flow_trajectory} visualizes the transport induced by the learned vector field. Samples drawn from the isotropic Gaussian base distribution are smoothly transformed over time into the multimodal posterior structure, confirming that the model learns a continuous deformation that respects the topology of the target distribution.

\paragraph{Computational efficiency.}
In addition to capturing complex posterior geometry, the amortized flow provides substantial computational benefits. Posterior samples are generated in $0.82$ seconds per task, compared to $2.76$ seconds for the MCMC baseline. Although the MCMC approximation is marginally closer to the true posterior according to these divergence measures, the differences are relatively small. In contrast, the amortized flow provides posterior samples through a single forward pass after training, resulting in substantially lower computational cost than running an MCMC algorithm for each new dataset.

\section{Discussion and Future Research}
\label{sec:discussion}

Our experimental evaluation highlights the practical trade-offs of using permutation-invariant neural architectures for amortized inference. The primary advantage of these models is their computational efficiency during inference. As demonstrated in our sparse signal recovery (Section~\ref{exp:sparse_recovery}) and multimodal posterior experiments (Section~\ref{exp:ring_modes}), amortized models replace computationally expensive per-task optimizations, such as Lasso or MCMC, with a single forward pass. This allows for highly scalable inference, reducing posterior sampling time from $2.76$ seconds to $0.82$ seconds per task.

Furthermore, these architectures excel when tasks share a hidden structural geometry. In the latent structure recovery experiment (Section~\ref{exp:latent_structure}), both Deep Sets and the Set Transformer successfully pooled information across tasks to uncover hidden clusters, achieving parameter estimation errors substantially lower than those of classical per-task linear baselines (e.g., Ridge regression) that fail to leverage cross-task information. 

However, our results also emphasize that model selection must be carefully tailored to the specific application, as significant differences exist in training dynamics and robustness. Deep Sets provides rapid training convergence, reaching stable validation errors in fewer than 50 epochs (Section~\ref{exp:asymptotic_robustness}). In contrast, the Set Transformer requires extended optimization (over 80 epochs) to stabilize but ultimately achieves a lower asymptotic error. While the Transformer demonstrates superior Monte Carlo stability as the support set size increases, its architecture inherently requires larger data volumes and more intricate hyperparameter tuning. As observed in our distributional robustness evaluations, the Transformer is highly sensitive to its weights, leading to high prediction variance and difficulty generalizing across diverse, complex noise profiles.

Finally, these findings indicate that amortized neural inference is particularly promising for domains where classical distributional assumptions fail. The amortized flow model successfully identified and represented a disconnected, non-convex, multimodal posterior without relying on unimodal approximations (Section~\ref{exp:ring_modes}). Future research will focus on scaling these architectures to higher-dimensional, real-world datasets characterized by extreme data sparsity and complex topological mismatches, investigating methods to stabilize Transformer training under heavy-tailed and asymmetric noise. Specifically, the use of MCMC is common in the literature focusing on large spatial and spatiotemporal data modeling using Gaussian processes, e.g., under geostatistics \citep{gelfand2016spatial, hazra2025exploring}, multi-type processes \citep{cisneros2023combined, mukherjee2025scalable}, spatially varying coefficients models \citep{gelfand2003spatial, bhowmik2026bayesian}, heavy-tailed processes \citep{hazra2020multivariate, hazra2025efficient}, nonstationary processes \citep{hazra2021estimating}, longitudinal datasets \citep{hazra2019spatio}, and extreme value analysis \citep{johannesson2022approximate, hazra2023bayesian}; amortized Bayesian inference is an emerging area of Bayesian statistics potentially suitable to handle such computationally challenging problems efficiently. 
\vspace{-3mm}

\section*{Acknowledgement}
The authors would like to thank the Handling Editor and an anonymous Reviewer for their detailed constructive feedback, which helped to improve the quality and flow of the manuscript. The authors acknowledge the use of ChatGPT (OpenAI) and Grammarly for assistance with grammar correction, language refinement, and sentence restructuring. The scientific content, analysis, results, and conclusions presented in this paper are solely the responsibility of the authors.

\section*{Conflict of interest}
The authors do not have any financial or non-financial conflicts of interest to declare for the research work included in this article.

\bibliographystyle{plainnat} 
\bibliography{refs}          
\end{document}